\documentclass[journal]{IEEEtran}
\ifCLASSINFOpdf
\else
\fi

\usepackage{amsfonts}
\usepackage{amsmath}
\usepackage{amsthm}
\usepackage{graphicx}
\usepackage{algorithm} 
\usepackage{algorithmic} 
\usepackage{multirow}
\usepackage{booktabs}
\usepackage{footmisc}

\usepackage{flushend}

\newtheorem{theorem}{Theorem}

\begin{document}
%

\title{Cost-Sensitive Feature Selection\\ by Optimizing F-measures}

\author{Meng Liu, Chang Xu, Yong Luo, Chao Xu,~\IEEEmembership{Member,~IEEE}  \\ Yonggang Wen,~\IEEEmembership{Senior Member,~IEEE} and Dacheng Tao,~\IEEEmembership{Fellow,~IEEE,}
	\thanks{Meng Liu and Chao Xu are with the Key Laboratory of Machine Perception (Ministry of Education), Cooperative Medianet Innovation Center, School of Electronics Engineering and Computer Science, Peking University, Beijing, 100871, China. (Email: mengliu@pku.edu.cn, xuchao@cis.pku.edu.cn)}
	\thanks{Chang Xu and Dacheng Tao are with the UBTECH Sydney Artificial Intelligence Centre and the School of Information Technologies in the Faculty of Engineering and Information Technologies at University of Sydney, 6 Cleveland St, Darlington, NSW 2008, Australia. (Email: c.xu@sydney.edu.au, dacheng.tao@sydney.edu.au)  }
	\thanks{Yong Luo and Yonggang Wen are with the School of Computer Science and Engineering, Nanyang Technological University, 639798, Singapore. (Email: yluo180@gmail.com, ygwen@ntu.edu.sg) }
	\thanks{\textcopyright 20XX IEEE. Personal use of this material is permitted. Permission from IEEE must be obtained for all other uses, in any current or future media, including
reprinting/republishing this material for advertising or promotional purposes, creating new collective works, for resale or redistribution to servers or lists, or reuse of any copyrighted component of this work in other works.}
}

%
%

\markboth{IEEE TIP-15403-2016 Final Version,~Vol.~XX, No.~X, December~2017}%
{Meng Liu \MakeLowercase{\textit{et al.}}: Cost-Sensitive Feature Selection by Optimizing F-measures}

\maketitle

\begin{abstract}

Feature selection is beneficial for improving the performance of general machine learning tasks by extracting an informative subset from the high-dimensional features. Conventional feature selection methods usually ignore the class imbalance problem, thus the selected features will be biased towards the majority class. Considering that F-measure is a more reasonable performance measure than accuracy for imbalanced data, this paper presents an effective feature selection algorithm that explores the class imbalance issue by optimizing F-measures. Since F-measure optimization can be decomposed into a series of cost-sensitive classification problems, we investigate the cost-sensitive feature selection (CSFS) by generating and assigning different costs to each class with rigorous theory guidance. After solving a series of cost-sensitive feature selection problems, features corresponding to the best F-measure will be selected. In this way, the selected features will fully represent the properties of all classes. Experimental results on popular benchmarks and challenging real-world datasets demonstrate the significance of cost-sensitive feature selection for the imbalanced data setting and validate the effectiveness of the proposed method.

\end{abstract}


\IEEEpeerreviewmaketitle

\section{Introduction}

\IEEEPARstart{F}{eature} selection aims to select a small subset of features from the input high-dimensional features by reducing noise and redundancy to maximize relevance to the target, such as class labels in image classification \cite{TIP2015CovertClassification,TIP2016multimodalClassification,Wang2015Feature,Nie2008Trace,luo2014decomposition}. It can improve the generalization ability of learning models, reduce the computational complexity, and thus benefit the subsequent learning tasks. Considering these advantages, a large number of works have been done in this literature \cite{CVPR2012Boosting,CVPR2012NMF,TIP2015UnSuperFL,IJCAI2011YangL21,IJCAI2015Geometric,IJCAI2015feature,TIP2006similarity,luo2016large,Tao2016Effective}. In general, these methods can be classified into three categories: (1) Filter methods that evaluate features by only examining the characteristics of data. ReliefF \cite{reliefF}, mRMR \cite{mRMR}, F-statistic \cite{F-statistic} and Information Gain \cite{InfoGain} are among the most representative algorithms. (2) Wrapper methods, which use the prediction method as a black box to score the feature subsets, such as correlation-based feature selection (CFS) \cite{CFS} and support vector machine recursive feature elimination (SVM-RFE) \cite{SVMRFE}. (3) Embedded methods, which directly incorporate the feature selection procedure into the model training process, such as regularized regression-based feature selection methods \cite{nips2010rfs,cvpr2015ufs}. These methods introduce group sparse regularization into the optimization of the regression model and often have low computational complexity.

Although all these aforementioned feature selection algorithms are promising and effective, most of them have been developed by implicitly or explicitly assuming the ideal data sampling, where the numbers of samples for all classes have no significant difference. However, in practice, the class imbalance issue, \textit{i.e.}, the minority class usually includes much fewer examples than the majority class does, is common in machine learning and pattern recognition. Specifically, the class imbalance of real-world datasets consists of two aspects: statistical distribution and algorithmic application. In the test set of handwritten digit dataset USPS \cite{USPSdataset}, the proportion of digit-0 is 18\%, while digit-7 is 7\%. Moreover, we often transform a multi-class/multi-label classification problem into multiple binary classification problems \textit{one-vs-all} strategy, where negative samples are much more than positive samples in each binary classification problem \cite{TIP2016multimodalClassification,SIGKDD2008fast,TIP2015spatialClass,SVMRFE,TIP2015cbirFExtr}. This encourages conventional feature selection methods to select the features describing the majority classes rather than those features representing the minority class. Given the subsequent classification or clustering task, it is then difficult to derive an optimal solution based on the biased selected features.

Furthermore, most classifier-dependent feature selection methods \cite{nips2010rfs,cvpr2015ufs,rebuttle2006combining,IJCAI2013Robust,IJCAI2011joint} are facing the same problem. These methods treat the misclassification costs of different classes equally and choose the feature subset with the highest classification accuracy, which is not a satisfying performance measure for the imbalanced training data. In this situation, classifiers are usually overwhelmed by the majority class due to the neglect of the minority class. These methods can thus be called as cost-blind feature selection methods \cite{PAMI2010cost}.


Since F-measure favors high and balanced values of precision and recall \cite{nips2014optimizing}, it is a more reasonable performance measure than accuracy for the class imbalance situation \cite{nips2014optimizing,FmExact,FmMLC}. F-measure has been widely used for evaluating the performance of binary classification, and its variants for multi-class and multi-label classification have been studied recently \cite{FmTopical,FmPlugin,FmExact,FmTaleTwo,FmMLC}. Various methods of optimizing F-measures have been proposed in the literature, and generally they fall into two paradigms. The decision-theoretic approach (DTA) \cite{DTA} first estimates a probability model, and then computes the optimal predictions according to the model. On the other hand, empirical utility maximization (EUM) approach \cite{jansche2005maximum,Joachims2005large} follows a structured risk minimization principle. Optimizing F-measure directly is often difficult since the F-measure is non-convex, thus approximation methods are often used instead, such as the algorithms for maximizing a convex lower bound of F-measure for support vector machines \cite{Joachims2005large}, and maximizing the expected F-measure of a probabilistic classifier using a logistic regression model \cite{jansche2005maximum}. A simpler method is to threshold the scores obtained by classifiers to maximize the empirical F-measure. Though simple, this method has been found effective and is commonly applied \cite{nips2014optimizing,yang2001study}. Recent works \cite{nips2014optimizing,FmTaleTwo,koyejo2014consistent,narasimhan2014statistical} study the pseudo-linear property of F-measures where they are functions of per-class false negative/false positive rate. In these scenarios, F-measure optimization can be reduced to a series of cost-sensitive classification problems, and an approximately optimal classifier for the F-measure is guaranteed.

Inspired by the recent successes on F-measure optimization \cite{nips2014optimizing,FmTaleTwo}, we propose a novel approach, called cost-sensitive feature selection (CSFS), to optimize F-measures of the classifiers used in embedded feature selection \cite{nips2010rfs,cvpr2014matrix,NCFS} to overcome the class imbalance problem as mentioned. Through the reduction of F-measure optimization to cost-sensitive classification, the classifiers of regularized regression-based feature selection methods are modified into cost-sensitive. By solving a series of cost-sensitive feature selection problems, features will be selected according to the optimal classifier with the largest F-measure. Therefore, the class imbalance is taken into consideration, and the selected features will fully represent both majority class and minority class. We apply our method to both multi-class and multi-label tasks. Promising results on several benchmark datasets as well as some challenging real-world datasets have validated the efficiency of the proposed method.

The main contributions of our work are as follows.

\begin{itemize}
	\item We propose that F-measure can be used to measure the performance of feature selection methods since F-measure is more reasonable than accuracy for imbalanced data classification. Although some previous works \cite{SIGKDD2008fast,KDE2010combating} have been proposed to deal with the feature selection and class imbalance problems, our method is one of the first works that handle the class-imbalance problem in feature selection by optimizing F-measure.
	\item We modify the traditional feature selection methods into cost-sensitive by designing and assigning different costs to each class. Based on the general reduction framework of F-measure optimization, our method can obtain an optimal solution.
\end{itemize}

 The rest of this paper is organized as follows. Section II introduces the notations and summarizes the preliminary knowledge on the reduction of F-measure optimization to cost-sensitive classification. The description and formulation of the algorithm are demonstrated in Section III. Extensive  experimental results conducted on the synthetic data, binary-class,  multi-class and multi-label datasets are reported in Section IV. We conclude our work in Section V.

%
%
%
%
%
%
%
%

\section{Notations and Preliminary Knowledge}

In this section, we introduce the notations used in this paper, and provide a brief introduction to F-measure optimization reduction.

\subsection{Notations}

In this paper, we present matrices as bold uppercase letters and vectors as bold lowercase letters. Given a matrix $\mathbf{W}=[w_{ij}]$, we denote $\mathbf{w}^i$ as its $i$-th row and $\mathbf{w}_j$ as its $j$-th column.

For $p>0$, the $\ell_p$-norm of the vector $\mathbf{b}\in\mathbb{R}^n$ is defined as $\|\mathbf{b}\|_p=(\sum_{i=1}^n|b_i|^p)^{\frac{1}{p}}$. The $\ell_{p,q}$-norm of the matrix $\mathbf{W}\in\mathbb{R}^{n\times m}$ is defined as $\|\mathbf{W}\|_{p,q}=(\sum_{i=1}^n\|\mathbf{w}^i\|_p^q)^{\frac{1}{q}}$, where $p>0$ and $q>0$. The symbol $\odot$ denotes the element-wise multiplication. The transpose, trace, inverse of a matrix $\mathbf{W}$ are denoted as $\mathbf{W}^T$, $tr(\mathbf{W})$ and $\mathbf{W}^{-1}$ respectively.

\subsection{F-Measure Optimization Reduction}

\begin{figure}[!tbp]
	\centering
	\includegraphics[width=0.47\textwidth]{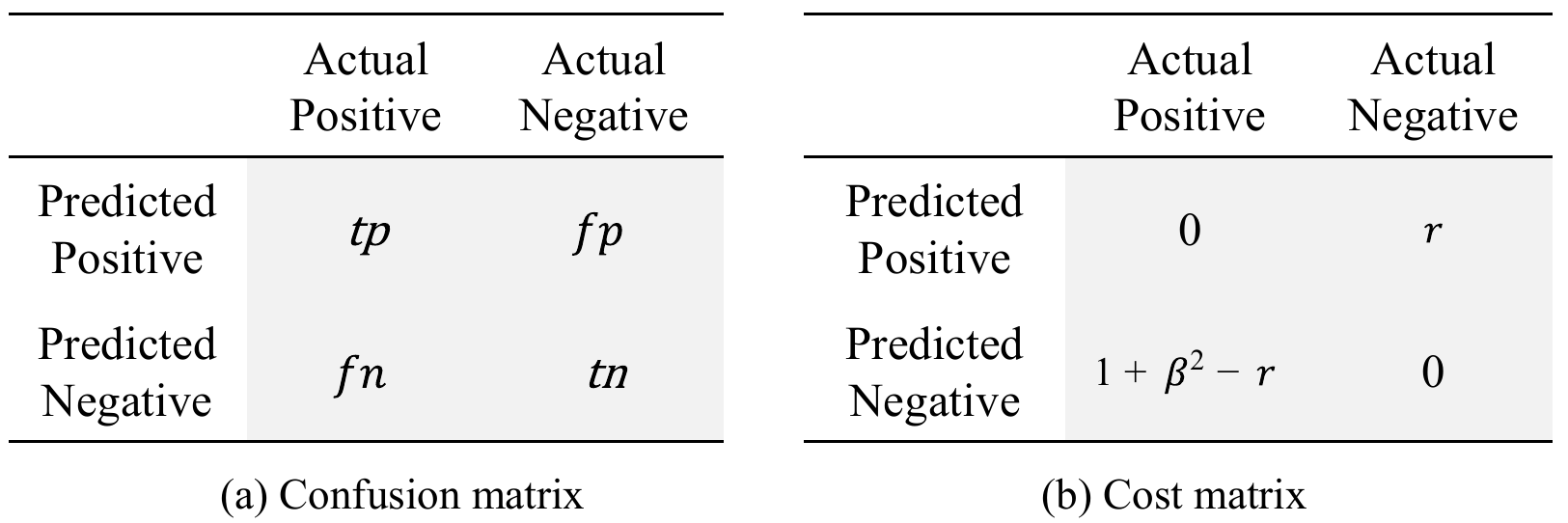}
	\caption{Confusion matrix and cost matrix in the context of binary classification. F$_\beta$-measure can be represented as the sum of the elements of the Hadamard product of confusion matrix and cost matrix \cite{nips2014optimizing}. }
	\label{fig:con_cost_table}
\end{figure}

Before presenting the formulation of the proposed method, we first summarize some preliminary knowledge on the reduction of F-measure optimization to cost-sensitive classification. There are four possible outcomes for a given binary classifier:  true positives $tp$, false positives $fp$, false negatives $fn$, and true negatives $tn$. They are represented as a confusion matrix in Figure \ref{fig:con_cost_table}(a).  F-measure can be defined in terms of the marginal probabilities of classes and the per-class false negative/false positive probabilities. The marginal probability of label $k$ is denoted by $P_k$, and the per-class false negative probability and false positive probability of a classifier $h$ are denoted by $FN_k(h)$ and $FP_k(h)$, respectively \cite{nips2014optimizing}. These probabilities of a classifier $h$ can be summarized by the error profile $\mathbf{e}(h)$:
\begin{equation}
\mathbf{e}(h)=(FN_1(h),FP_1(h),\dots,FN_m(h),FP_m(h)),
\end{equation}
where $m$ is the number of labels, $e_{2k-1}$ of $\mathbf{e}(h)\in\mathbb{R}^{2m}$ is the false negative probability of class $k$ and $e_{2k}$ is the false positive probability. In binary classification, we have $FN_2=FP_1$. Thus, for any $\beta>0$, F-measure can be written as a function of error profile $\mathbf{e}$:
\begin{equation}\label{eqn:fm_errorprofile}
F_{\beta}(\mathbf{e})=\frac{(1+\beta^2)(P_1-e_1)}{(1+\beta^2)P_1-e_1+e_2}.
\end{equation}

There are several definitions of F-measures in multi-class and multi-label classification. Specifically, we can transform the multi-class or multi-label classification into multiple binary classification problems, and the average over the F$_\beta$-measures of these binary problems is defined as the macro-F-measure. According to \cite{nips2014optimizing}, we can also define the following micro-F-measure $mlF_\beta$  for multi-label classification:
\begin{equation}\label{eqn:ml_microFm}
mlF_\beta(\mathbf{e})=\frac{(1+\beta^2)\sum_{k=1}^m(P_k-e_{2k-1})}{\sum_{k=1}^m((1+\beta^2)P_k+e_{2k}-e_{2k-1})}.
\end{equation}

Multi-class classification differs from multi-label classification in that a single class must be predicted for each example. According to \cite{microFm}, one definition of multi-class micro-F-measure, denoted as $mcF_\beta$ can be written as:
\begin{equation}\label{eqn:mc_microFm}
    mcF_\beta(\mathbf{e})=\frac{(1+\beta^2)(1-P_1-\sum_{k=2}^me_{2k-1})}
    {(1+\beta^2)(1-P_1)-\sum_{k=2}^me_{2k-1}+e_1}.
\end{equation}

We can notice that the fractional-linear F-measures presented in Eqs. (\ref{eqn:fm_errorprofile}-\ref{eqn:mc_microFm}) are pseudo-linear functions with respect to $\mathbf{e}$. The important property of pseudo-linear functions is that their level sets, as function of the false negative rate and the false positive rate, are linear. Based on this observation, a recent work \cite{nips2014optimizing} was proposed  for F-measure maximization by reducing it into cost-sensitive classification, and proved that the obtained optimal classifier for a cost-sensitive classification problem with label dependent costs is also an optimal classifier for F-measure. This method can be separated into three steps. Firstly, the F-measure range $[0,1]$ is discretized into a set of evenly spaced values $\{r_i\}$ without considering its variable property under label switching. 
Secondly, for each given F-measure value $r$,  cost function $\mathbf{a}:\mathbb{R}^1_+\rightarrow\mathbb{R}^{2m}_+$ generates a cost vector $\mathbf{a}(r)\in\mathbb{R}^{2m}$ and assigns costs to the elements of error profile $\mathbf{e}$. We will use binary-class $F_\beta(\mathbf{e})$ as an example to illustrate the procedure of generating costs. 

As a pseudo-linear function, the level sets of $F_\beta(\mathbf{e})$ are linear \cite{koyejo2014consistent,narasimhan2014statistical}. For a given discretized F-measure value $r$, by setting denominator of Eq. (\ref{eqn:fm_errorprofile}) strictly positive and $F_\beta(\mathbf{e})\le r$, we can get
	\begin{equation}\label{eqn_generatingBinaryCosts}
	(1+\beta^2-r)e_1+re_2+(1+\beta^2)P_1(r-1)\ge 0.
	\end{equation}
The non-negative coefficients $1+\beta^2-r$ and $r$ can be interpreted as the costs of $e_1$ and $e_2$. Thus, the cost function $\mathbf{a}^{F_\beta}(r)$ for binary-class $F_\beta$ can be represented as follows:
\begin{equation}\label{eqn_costsbinary}
	a^{F_\beta}_i(r) = \left\{
	\begin{array}{ll}
	1+\beta^2-r &\text{if}~i=1 \\
	r &\text{if}~i=2\\
	0 & \text{otherwise.}
	\end{array}
	\right.
\end{equation}
These costs are shown as a cost matrix in Figure \ref{fig:con_cost_table}(b) \cite{nips2014optimizing}. Therefore, the goal of optimization is changed to minimize the total cost $\langle \mathbf{a}(r),\mathbf{e}(h)\rangle$, which is the inner product of cost vector and error profile \cite{nips2014optimizing}.

\begin{figure}[!tbp]
	\centering
	\includegraphics[width=0.45\textwidth]{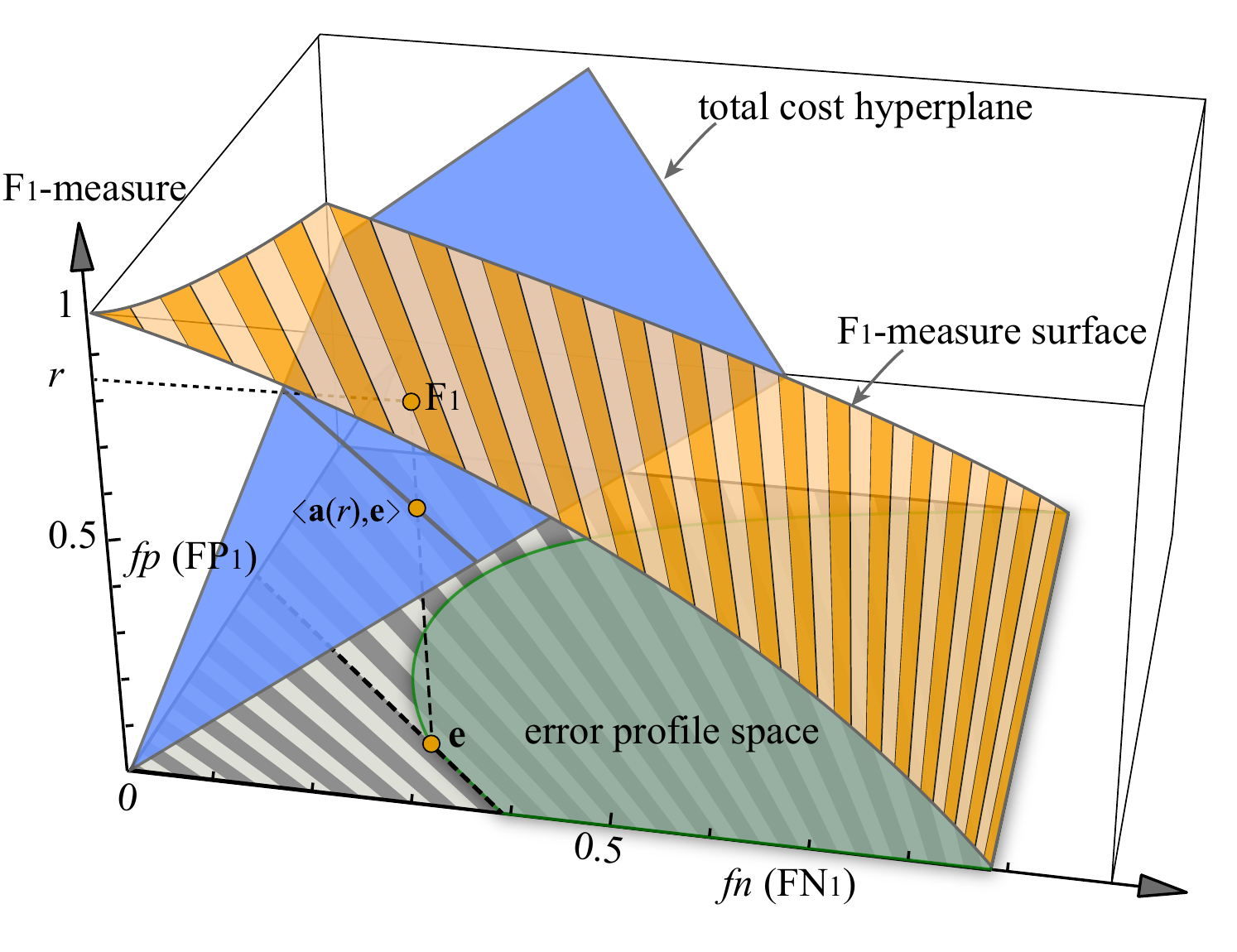}\\
	\caption{Illustration of F$_1$-measure surface with level sets and total cost hyperplane for a given cost vector in the context of binary classification. F$_1$-measure is a nonlinear function with respect to \emph{fn} and \emph{fp}. When cost vector $\mathbf{a}(r)$ is fixed, the total cost hyperplane is a linear function of \emph{fn} and \emph{fp}. Error profile space contains all possible values of $\mathbf{e}$ illustratively. Intuitively, we can notice that higher values of F$_1$-measure entail lower values of total cost $\langle \mathbf{a}(r),\mathbf{e}\rangle$. }
	\label{fig:surfaces}
\end{figure}

Similarly, the cost function $\mathbf{a}^{mlF_\beta}(r)$ of multi-label micro-F-measure $mlF_\beta$ can be represented as follows:
\begin{equation}\label{eqn_costsMLF}
a^{mlF_\beta}_i(r) = \left\{
\begin{array}{ll}
1+\beta^2-r &\text{if}~i~\text{is odd} \\
r &\text{if}~i~\text{is even}
\end{array}
\right.
\end{equation}
and the cost function $\mathbf{a}^{mcF_\beta}(r)$ of multi-class micro-F-measure $mcF_\beta$ is:
\begin{equation}\label{eqn_costsMCF}
a^{mcF_\beta}_i(r) = \left\{
\begin{array}{ll}
r &\text{if}~i=1\\
1+\beta^2-r &\text{if}~i~\text{is odd and}~i \neq 1 \\
0 & \text{otherwise.}
\end{array}
\right.
\end{equation}
Lastly, cost-sensitive classifiers for each $\mathbf{a}(r)$ are learned to minimize the total cost $\langle \mathbf{a}(r),\mathbf{e}(h)\rangle$, and the one with largest F-measure on the validation set is selected as the optimal classifier. Figure \ref{fig:surfaces} shows that the higher the F-measure value, the lower the total cost. This indicates that maximizing F-measure can be achieved by minimizing the corresponding total cost.

The determination of misclassification costs usually requires experts or prior knowledge. These costs  reflect how severe one kind of mistake compared with other kinds of mistakes \cite{PAMI2010cost}, which will result in numerous cost matrices based on different cost ratios. By contrast, the costs generated from the F-measure optimization reduction algorithm have explicit value ranges once the discretized step is determined.

\begin{figure*}[!htbp]
\centering
\includegraphics[width=0.98\textwidth]{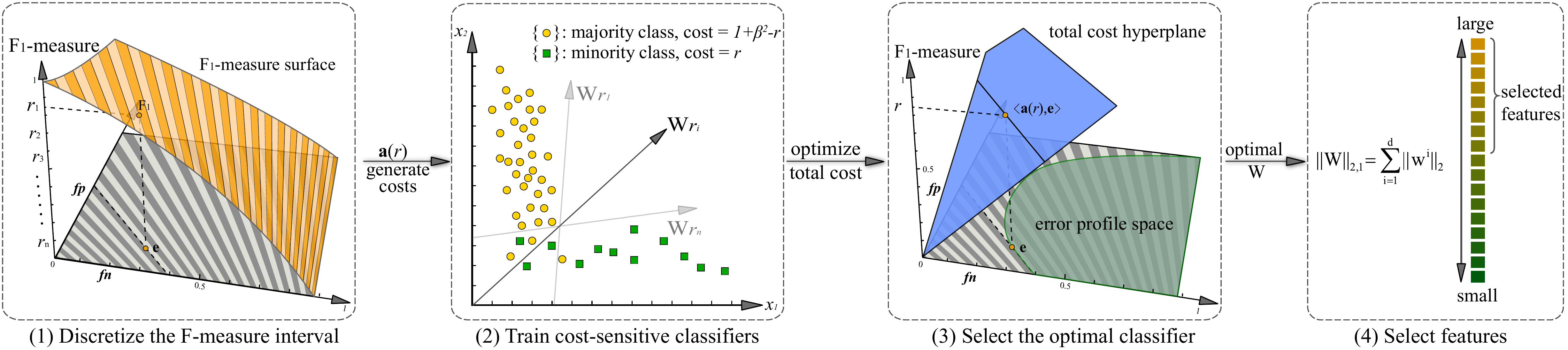}\\
\caption{System diagram of the proposed cost-sensitive feature selection (CSFS) model in the case of binary classification. This model can be divided into four stages. (1) Discretize the F-measure interval  to obtain a set of evenly spaced values $\{r_1,\dots,r_n\}$. (2) For a given $r_i$, cost function $\mathbf{a}(r_i)$ generates costs $1+\beta^2-r_i$ for the false negative and $r_i$ for the false positive, thus we can get a series of cost-sensitive classifiers. (3) Select the optimal classifier with the largest F-measure value on the validation set. (4) Select the top-ranking features according to the projection matrix $\mathbf{W}$ of the optimal classifier by sorting $\|\mathbf{w}^i\|$ $(1\leq i\leq d)$ in descending order.
}
\label{fig:mainIllustration}
\end{figure*}

\section{Cost-Sensitive Feature Selection}

Inspired from the reduction of F-measure optimization to cost-sensitive classification, we modify the classifiers used in regularized regression-based feature selection methods into cost-sensitive by adding properly generated costs, and select features according to the classifier with the optimal F-measure performance. This leads to a novel cost-sensitive feature selection (CSFS) method by optimizing F-measures. A schematic illustration of the proposed method is shown in Figure \ref{fig:mainIllustration}.

\subsection{Problem Formulation}

Given training data, let $\mathbf{X}=[\mathbf{x}_1,\dots,\mathbf{x}_n]\in\mathbb{R}^{d\times n}$ denote feature matrix with $n$ samples and the feature dimension is $d$. The corresponding label matrix is given by $\mathbf{Y}=[\mathbf{y}^1;\dots;\mathbf{y}^n]\in\{-1,1\}^{n\times m}$ where $\mathbf{y}^i$ is a row vector of the labels for the $i$-th example, and $m$ is the number of class labels.  The general formulation of regularized regression-based feature selection methods \cite{nips2010rfs,cvpr2015ufs,Xiang2012Discriminative}, which aim to obtain a projection matrix $\mathbf{W}\in\mathbb{R}^{d\times m}$, can be summarized as follows:
\begin{equation}
\min_{\mathbf{W}} \mathcal{L}(\mathbf{X}^T\mathbf{W} - \mathbf{Y})+\lambda\mathcal{R}(\mathbf{W}),
\label{eqn:Loss_L21}
\end{equation}
where $\mathcal{L}(\cdot)$ is the norm-based loss function of the prediction residual, $\mathcal{R}(\cdot)$ is the regularizer that introduces sparsity to make $\mathbf{W}$ applicable for feature selection, and $\lambda$ is a trade-off parameter. For simplicity, the bias has been absorbed into $\mathbf{W}$ by adding a constant value 1 to the feature vector of each example. Such methods have been widely used in both multi-class and multi-label data situations \cite{nips2010rfs,NCFS,cvpr2015ufs,cvpr2014matrix}. However, they are designed to maximize the classification accuracy, which is unsuitable for highly imbalanced classes situations \cite{nips2014optimizing}, since equal costs are assigned to different classes.

To deal with this problem, we now present a new feature selection method, which optimizes F-measure by modifying the classifiers of regularized regression-based feature selection into cost-sensitive.  Without loss of generality, we start with the illustration on the cost-sensitive feature selection under the binary-class setting, where the label vector is $[y_1;y_2;\dots;y_n]\in\{-1,1\}^{n \times 1}$. As mentioned previously, the misclassification cost for positive class is $1+\beta^2-r$ and the misclassification cost for negative class is $r$. Thus for each class, we obtain a cost vector $\mathbf{c}=[c_1,\dots,c_n]^T\in\mathbb{R}^n$, where $c_i=1+\beta^2-r$ if $y_i=1$, and $c_i=r$ if $y_i=-1$. The formulation of total cost for all samples can be given as follows:
\begin{equation}
    \label{eqn:cs_binary}
    \min_{\mathbf{w}}\sum_{i=1}^n\mathcal{L}((\mathbf{x}_i^T\mathbf{w}- y_i)\cdot c_i ) +\lambda\mathcal{R}(\mathbf{w}),
\end{equation}
where $\mathbf{w}\in\mathbb{R}^{d\times 1}$ is the projection vector. In multi-class and multi-label scenarios, the cost vector $\mathbf{c}_i\in\mathbb{R}^n$ for the $i$-th class can be obtained according to their per-class false negative/false positive cost generated by corresponding cost function $\mathbf{a}(r)$.  Denoting the cost matrix as $\mathbf{C}=[\mathbf{c}_1,\mathbf{c}_2,\dots,\mathbf{c}_m]\in\mathbb{R}^{n\times m}$, we obtain the following formulation:
\begin{equation}
    \label{eqn:cs_multilabel}
    \min_{\mathbf{W}}\sum_{i=1}^n
    \mathcal{L}((\mathbf{x}_i^T\mathbf{W}-\mathbf{y}^i)\odot \mathbf{c}^i)
    +\lambda\mathcal{R}(\mathbf{W}),
\end{equation}
where $\mathbf{c}^i$ is the $i$-th row of $\mathbf{C}$ corresponding to the $i$-th example. Due to the rotational invariant property and robustness to outliers \cite{nips2010rfs}, we adopt $\ell_2$-norm based loss function as the specific form of $\mathcal{L}(\cdot)$ and the optimization problem becomes:
\begin{equation}
\min_{\mathbf{W}}\sum_{i=1}^n
\|(\mathbf{x}_i^T\mathbf{W}-\mathbf{y}^i)\odot \mathbf{c}^i\|_2
+\lambda\mathcal{R}(\mathbf{W}).
\end{equation}


By further considering that
\begin{equation}
    \sum_{i=1}^n\|(\mathbf{x}_i^T\mathbf{W}-\mathbf{y}^i)\odot \mathbf{c}^i\|_2
    =
    \|(\mathbf{X}^T\mathbf{W}-\mathbf{Y})\odot \mathbf{C}\|_{2,1},
\end{equation}
and taking the commonly used $\ell_{2,1}$-norm as regularization \cite{CVPR2012L21,cvpr2015ufs,nips2010rfs,cai2013exact}, we obtain the following compact form of the cost-sensitive feature selection (CSFS) optimization problem:
                            \begin{equation}\label{eqn:main_obj_L21}
    \min_{\mathbf{W}}\|(\mathbf{X}^T\mathbf{W}-\mathbf{Y})\odot \mathbf{C}\|_{2,1}+\lambda\|\mathbf{W}\|_{2,1}.
\end{equation}

As shown in Figure \ref{fig:mainIllustration}, we can get a series of cost-sensitive feature selection problems with different cost matrix $\mathbf{C}$ corresponding to each F-measure value $r$. After obtaining the optimal $\mathbf{W}$, features can be selected by sorting $\|\mathbf{w}^i\|$ $(1\leq i\leq d)$ in descending order. If  $\|\mathbf{w}^i\|$ shrinks to zero, the $i$-th feature is less important and will not be selected.

It is worth mentioning that the loss and regularization of Eq. (\ref{eqn:main_obj_L21}) are not necessarily $\ell_{2,1}$-norm based terms. Other regularizations that have the effects of feature selection, such as ridge regularization and LASSO regularization \cite{nips2010rfs}, can also be applied to obtain the concrete form of objective function.

\subsection{Optimization Method}

The difficulty of optimization for Eq. (\ref{eqn:main_obj_L21}) lies in the $\ell_{2,1}$-norm imposed on both loss term and regularization term, which is difficult to get an explicit solution. For a given F-measure $r$, the corresponding cost matrix $\mathbf{C}$ is fixed and thus $\mathbf{W}$ is the only variable in Eq. (\ref{eqn:main_obj_L21}). Thus, we develop an iterative algorithm to solve this problem. Taking the derivative of the objective function with respect to $\mathbf{w}_k(1\leq k\leq m)$ and setting it to zero, we obtain:
\begin{equation}
\mathbf{X} \mathbf{U}_k\mathbf{G}\mathbf{U}_k \mathbf{X}^T\mathbf{w}_{k}
-\mathbf{X} \mathbf{U}_k\mathbf{G}\mathbf{U}_k \mathbf{y}_k + \lambda \mathbf{D}\mathbf{w}_{k} =0,
\end{equation}
where diagonal matrix $\mathbf{U}_k=diag(\mathbf{c}_k)$, $\mathbf{D}$ is a diagonal matrix with the $i$-th diagonal element as\footnote{When $\|\mathbf{w}^i\|_2=0$, Eq. (\ref{eqn:main_obj_L21}) is not differnetiable. Following \cite{nips2010rfs}, this problem can be solved by introducing a small perturbation to regularize $d_{ii}$ as $\frac{1}{2\sqrt{\|\mathbf{w}^i\|_2^2+\zeta}}$. Similarly, the $i$-th diagonal element $g_{ii}$ of $\mathbf{G}$ can be regularized as $\frac{1}{2\sqrt{\|((\mathbf{X}^T\mathbf{W}-\mathbf{Y})\odot \mathbf{C})^i\|_2^2+\zeta}}$. It can be verified that the derived algorithm minimizes the following problem: $\sum_{i=1}^n\sqrt{\|((\mathbf{X}^T\mathbf{W}-\mathbf{Y})\odot \mathbf{C})^i\|_2^2+\zeta}+ \lambda\sum_{i=1}^d\sqrt{\|\mathbf{w}^i\|_2^2+\zeta}$, which is apparently reduced to Eq. (\ref{eqn:main_obj_L21}) when $\zeta\rightarrow 0$.}
\begin{equation}
d_{ii}=\frac{1}{2\|\mathbf{w}^i\|_2},
\end{equation}
and $\mathbf{G}$ is a diagonal matrix with the $i$-th diagonal element as
\begin{equation}
g_{ii}=\displaystyle\frac{1}{2\|((\mathbf{X}^T\mathbf{W}-\mathbf{Y})\odot \mathbf{C})^i\|_2}.
\end{equation}
Each $\mathbf{w}_{k}$ can thus be solved in the closed form:
\begin{equation}
\label{eqn:each_w}
\mathbf{w}_{k}=
\displaystyle(\lambda \mathbf{D}+\mathbf{X} \mathbf{U}_k\mathbf{G}\mathbf{U}_k \mathbf{X}^T )^{-1}
(\mathbf{X} \mathbf{U}_k\mathbf{G}\mathbf{U}_k )\mathbf{y}_k.
\end{equation}

Since the solution of $\mathbf{W}$ is dependent on $\mathbf{D}$ and $\mathbf{G}$, we develop an iterative algorithm to obtain the ideal $\mathbf{D}$ and $\mathbf{G}$. The whole optimization procedure is described in Algorithm \ref{alg:CSFS}. In each iteration, $\mathbf{D}$ and $\mathbf{G}$ are calculated with current $\mathbf{W}$, and then each column vector $\mathbf{w}_k$ of $\mathbf{W}$ is updated based on the newly solved $\mathbf{D}$ and $\mathbf{G}$. The iteration procedure is repeated until the convergence criterion is reached.

\begin{algorithm}[!tbp] 
	\renewcommand{\algorithmicrequire}{\textbf{Input:}}
	\renewcommand\algorithmicensure {\textbf{Output:} }
	\caption{An iterative algorithm to solve the optimization problem in Eq. (\ref{eqn:main_obj_L21}).}
	\label{alg:CSFS}
	\begin{algorithmic}[1]
		\REQUIRE feature matrix $\mathbf{X}\in\mathbb{R}^{d\times n}$, label matrix $\mathbf{Y}\in\mathbb{R}^{n\times m}$ and discretized F-measure value $r$.
		\ENSURE projection matrix $\mathbf{W}\in\mathbb{R}^{d\times m}$.
		
		\STATE Generate cost function $\mathbf{a}(r)$ for binary-class, multi-label, multi-class tasks according to Eqs. (\ref{eqn_costsbinary}), (\ref{eqn_costsMLF}) and (\ref{eqn_costsMCF}), respectively.

		\STATE Calculate cost vector $\mathbf{c}_i\in\mathbb{R}^n(1\le i\le m)$  for the $i$-th class according to per-class FN$_i$/FP$_i$ cost in $\mathbf{a}(r)$.  
		\STATE Obtain cost matrix $\mathbf{C}=[\mathbf{c}_1,\mathbf{c}_2,\dots,\mathbf{c}_m]\in\mathbb{R}^{n\times m}$.
		\STATE Initialize $\mathbf{W}_0$ as a random matrix, $t=0$.
		\WHILE{not converging}
		\STATE \textbf{update} diagonal matrix  $\mathbf{D}_{t+1}$ where the $i$-th diagonal element is $\frac{1}{2\|\mathbf{w}_t^i\|_2}$.
		\STATE \textbf{update} diagonal matrix  $\mathbf{G}_{t+1}$ where the $i$-th diagonal element is $\frac{1}{2\|((\mathbf{X}^T\mathbf{W}_t-\mathbf{Y})\odot \mathbf{C})^i\|_2}$.
		\FOR{$k\leftarrow 1$ \textbf{to} $m$}
		\STATE $\mathbf{U}_k=diag(\mathbf{c}_k)$.
		\STATE $(\mathbf{w}_{t+1})_{k}=(\lambda \mathbf{D}_{t+1}+\mathbf{X} \mathbf{U}_k\mathbf{G}_{t+1}\mathbf{U}_k \mathbf{X}^T )^{-1}$

		\STATE $\qquad\qquad\quad\cdot(\mathbf{X} \mathbf{U}_k\mathbf{G}_{t+1}\mathbf{U}_k )\mathbf{y}_k$.

		\ENDFOR
		\STATE $t=t+1$.
		\ENDWHILE
		
	\end{algorithmic}
\end{algorithm}

\begin{figure}[!htbp]
	\centering
	\includegraphics[width=0.46\textwidth]{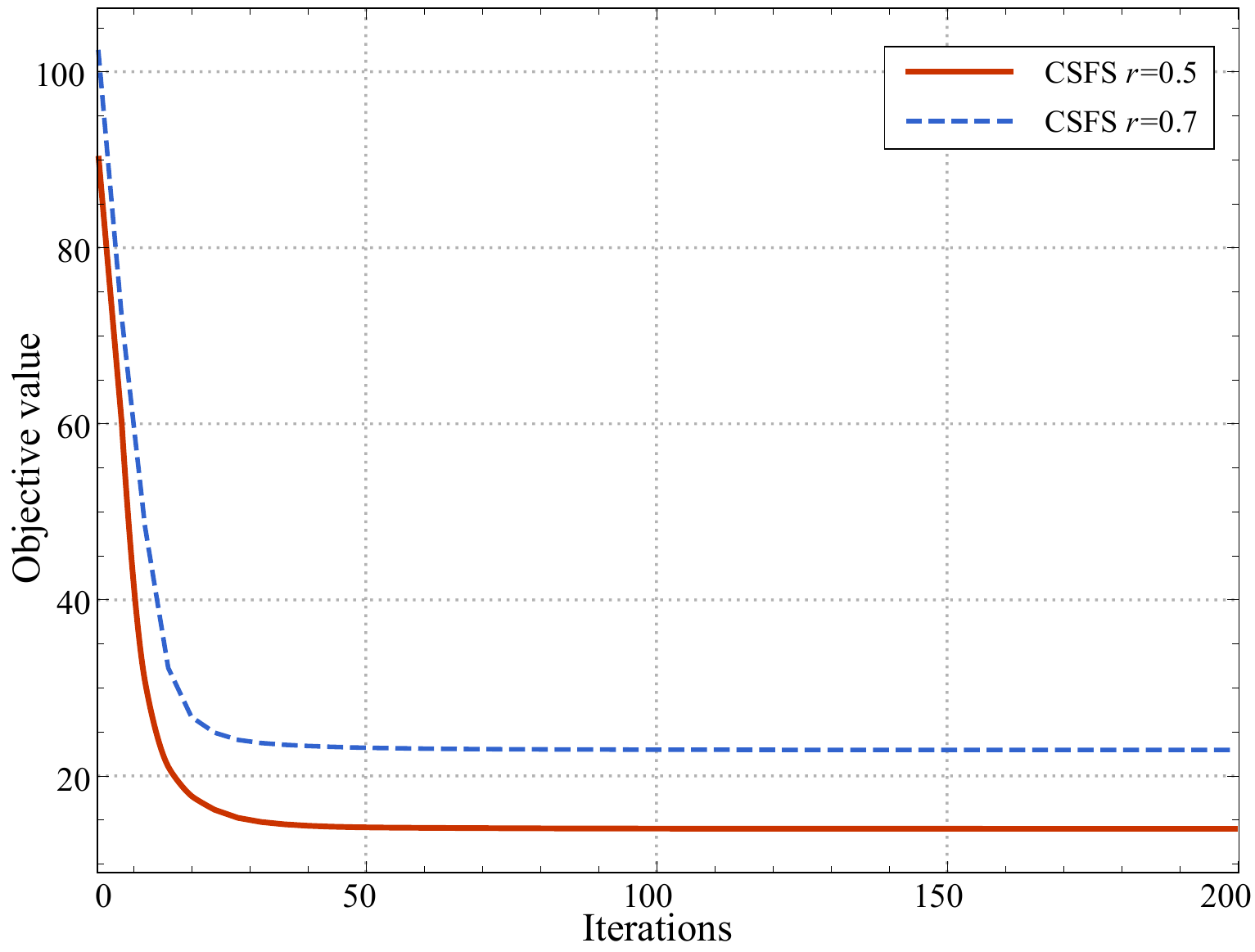}\\
	\caption{The convergence property of the proposed CSFS method. The two lines show how the objective values change along with the iterations given the discretized F-measure values $r=0.5$ and $r=0.7$, respectively.}
	\label{fig_IterObj}
\end{figure}

\subsection{Convergence Analysis}

The convergence of Algorithm \ref{alg:CSFS} is guaranteed by the following theorem:
\begin{theorem}\label{convergenceTheorem}
	Algorithm \ref{alg:CSFS} monotonically decreases the objective value of Eq. (\ref{eqn:main_obj_L21}) in each iteration,
	that is,
	\begin{equation}
	\begin{split}
	&\|(\mathbf{X}^T\mathbf{W}_{t+1}-\mathbf{Y})\odot \mathbf{C}\|_{2,1} +\lambda\|\mathbf{W}_{t+1}\|_{2,1}\leq\\
	&\|(\mathbf{X}^T\mathbf{W}_{t}-\mathbf{Y})\odot \mathbf{C}\|_{2,1} +\lambda\|\mathbf{W}_{t}\|_{2,1}.
	\end{split}
	\end{equation}
	
\end{theorem}

We put the convergence analysis of algorithm \ref{alg:CSFS} in the appendix. According to \cite{nips2010rfs}, the objective value of Eq. (\ref{eqn:main_obj_L21}) monotonically decreases in iterations, which implies that Algorithm \ref{alg:CSFS} can converge to a local minimum.


To demonstrate the convergence property of CSFS, we conduct an experiment on the USPS 1:10 subset (introduced in the following Section IV) to show in Figure \ref{fig_IterObj} the evolution of the objective value w.r.t. iterations. We can notice that the objective values of CSFS rapidly reach their steady state when the iteration number is around 40.

\subsection{Complexity Analysis}

In Algorithm \ref{alg:CSFS}, steps 6 and 7 calculate the diagonal elements which are computationally trivial, so the complexity mainly depends on the matrix multiplication and inversion in step 10. By using sparse matrix multiplication and avoiding dense intermediate matrices, the complexity of updating each $(\mathbf{w}_{t+1})_k$ is $O(d^2(n+d))$. Thus the complexity of the proposed algorithm is $O(Ttmd^2(n+d))$, where $t$ is the number of iterations, and $T$ is the number of discretized values of F-measure. Empirical results show that the convergence of Algorithm \ref{alg:CSFS} is rapid and $t$ is usually less than 50. Besides, $ T$ is usually less than 20. Therefore, the proposed algorithm is quite efficient. Since the update of each $(\mathbf{w}_{t+1})_k$ and the optimization for different costs are independent,  they can both be rapidly solved when sufficient computational resources are available and parallel computing is implemented. Thereby the whole cost-sensitive feature selection framework can be solved efficiently.


\section{Experiments}

In this section, we evaluate the performance of the proposed method with other feature selection methods. Specifically, the experiments consist of four parts. Firstly, we construct a two-dimensional binary-class synthetic dataset to show the influence of costs during feature selection process. Secondly, multiple binary datasets with different sampling ratios are extracted to validate the effectiveness of our CSFS in comparison to other multi-class feature selection methods. Lastly, extensive experiments are conducted on real-world multi-class and multi-label datasets. For multi-class classification, we use six multi-class datasets: one cancer dataset LUNG\footnote{http://www.ncbi.nlm.nih.gov/pubmed/11707567}, one object image dataset COIL20\footnote{\label{mcDatasets}http://www.cad.zju.edu.cn/home/dengcai/Data/MLData.html}, one spoken letter recognition dataset Isolet1\footref{mcDatasets}, one handwritten digit dataset USPS\footref{mcDatasets}, and two face image datasets YaleB\footref{mcDatasets} and UMIST\footnote{http://www.sheffield.ac.uk/eee/research/iel/research/face}. For multi-label classification, we use the Barcelona\footnote{http://mlg.ucd.ie/content/view/61}, MSVCv2\footnote{http://research.microsoft.com/en-us/projects/objectclassrecognition/} and TRECVID2005\footnote{http://www-nlpir.nist.gov/projects/tv2005/} datasets. We extract 384-dimensional color moment features on the Barcelona and MSRCv2 datasets, and 512-dimensional GIST features on TRECVID. For each dataset, we randomly select $1/3$ of the training samples for validation to tune the hyper-parameters. For datasets that do not have a separate test set, the data is first split to keep $1/4$ for testing. Information of multi-class and multi-label datasets is summarized in Table \ref{tbl:datasets}.

\begin{figure*}[!thbp]
	\centering
	\includegraphics[width=0.9\textwidth]{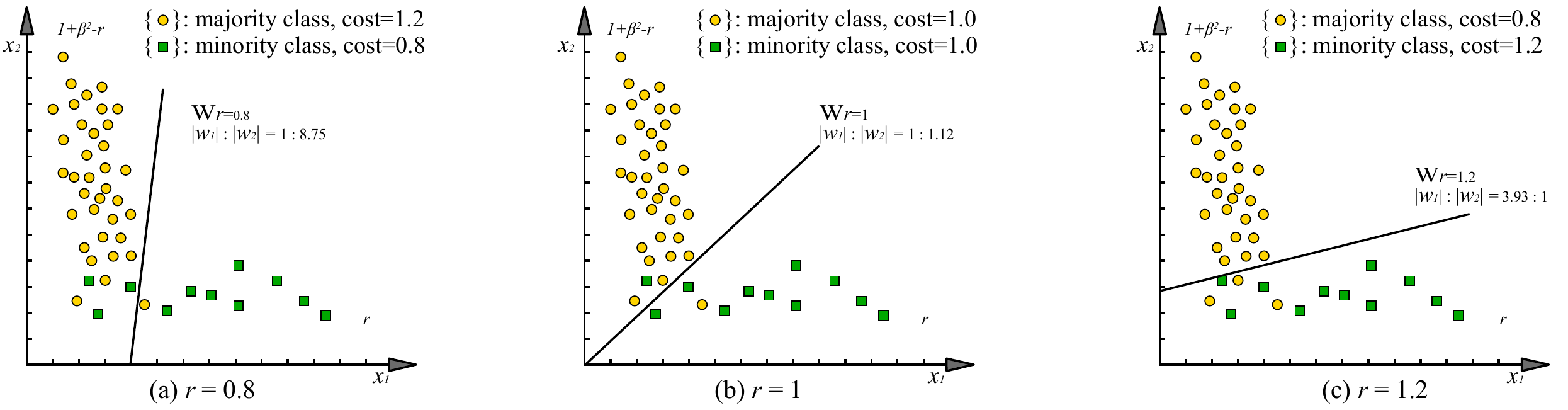}\\
	\caption{Illustration of how CSFS influences the features weights on a two-dimensional synthetic dataset. The feature weights, \textit{i.e.}, the absolute values of the coefficients of the projection vector $\mathbf{w}$ change with the assigned cost to each class. When discreted F-measure value $r=1.2$, the cost of minority class is larger than the cost of majority class, which makes the projection vector $\mathbf{w}$ bias towards feature $x_1$. In this case, the weight of feature of $x_1$ is larger than the weight of feature $x_2$, which is different from the situations when $r=0.8$ and $r=1.0$. }
	\label{fig:toyExp}
\end{figure*}

\begin{figure*}[!tbp]
	\centering
	\includegraphics[width=0.92\textwidth]{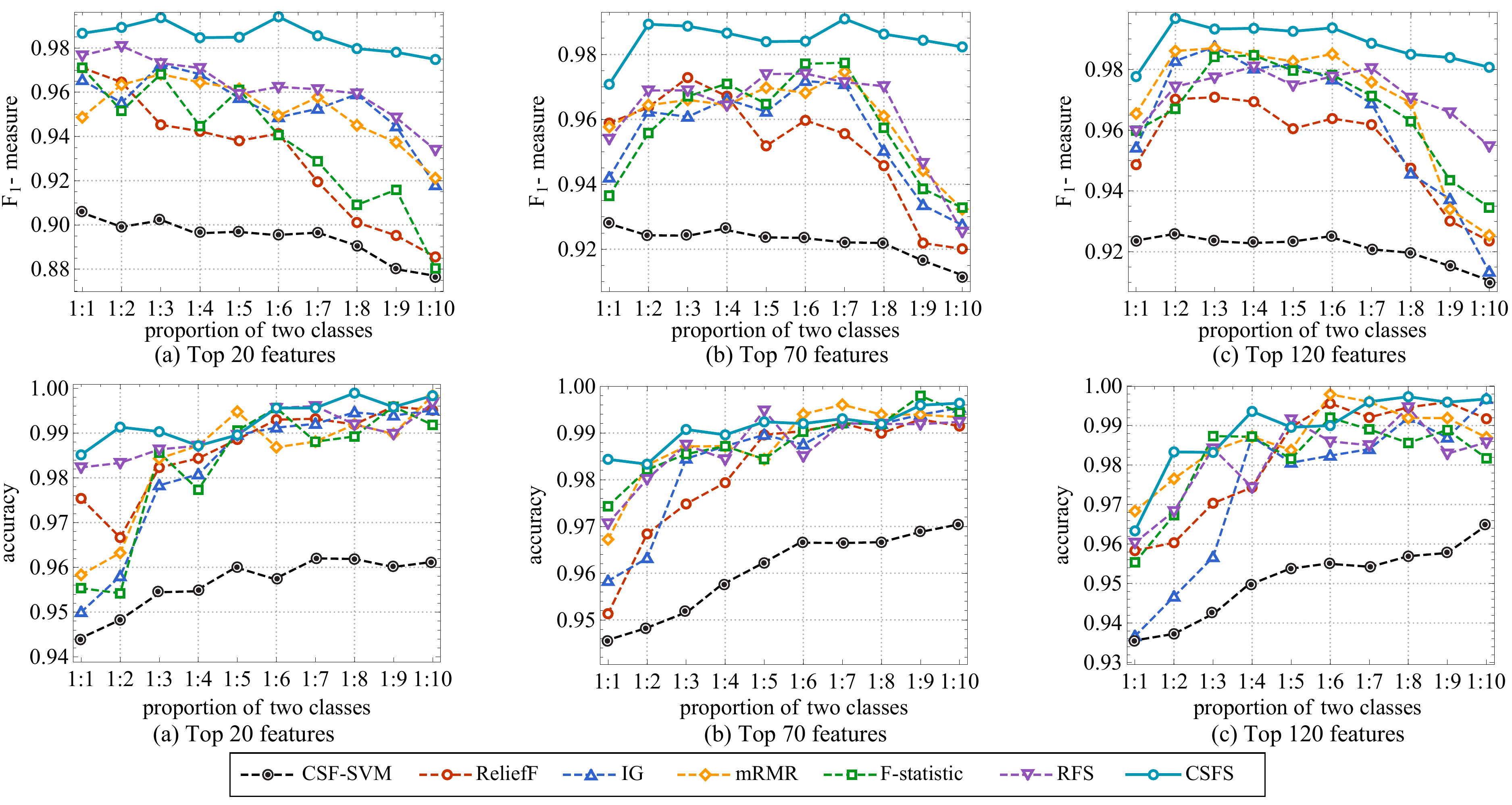}\\
	\caption{ Binary classification results using SVM in terms of F$_1$-measure and accuracy on ten binary datasets with different levels of imbalanced classes. The numbers of selected features are set as 20, 70 and 120. The sampling ratio for two classes varies from 1 to 10. For each fixed number of selected features, the chart above shows the F$_1$-measure and the chart below shows the accuracy of classification. CSF-SVM is a baseline method which first trains cost-sensitive SVMs to optimize F-measure and then performs feature selection by thresholding the weights.}
	\label{fig:imbalanceFmAcc}
\end{figure*}

\begin{figure*}[!thbp]
	\centering
	\includegraphics[width=0.92\textwidth]{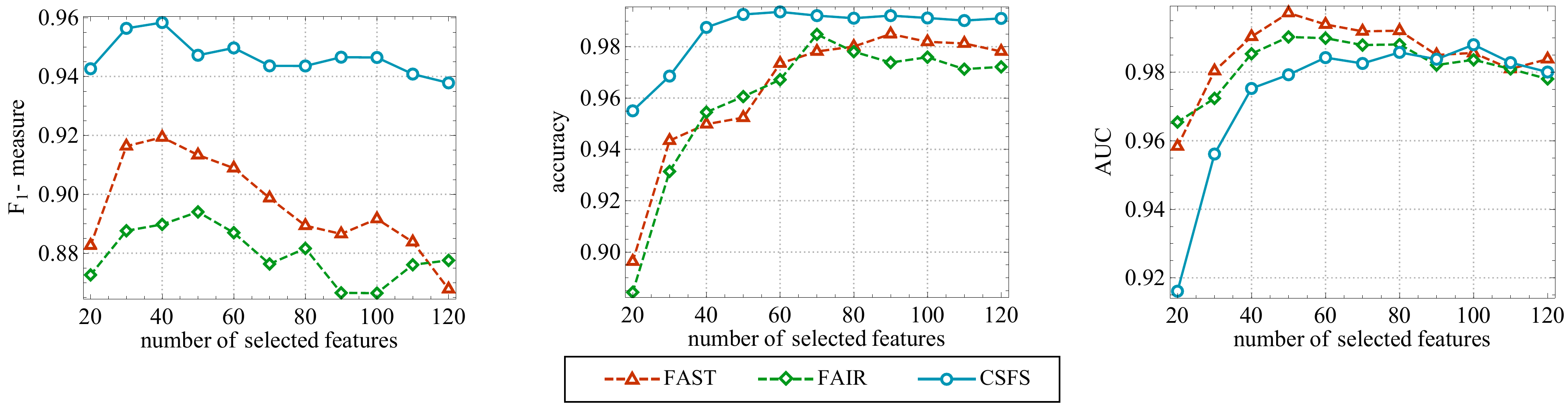}\\
	\caption{ Binary classification results using SVM in terms of F$_1$-measure, accuracy and AUC on the USPS 1:10 subset. FAST and FAIR are two binary-class feature selection methods which take the imbalance issue into consideration by optimizing AUC via sliding the thresholds.}
	\label{fig_binaryFastFairCsfs}
\end{figure*}

\begin{table}[!tbp]
	\setlength\tabcolsep{7pt}
	\renewcommand{\arraystretch}{1.1}
	\centering
	\caption{Summarization of the multi-class and multi-label datasets.}
	\begin{tabular}{|c|c|c|c|c|}
		\hline
		\multicolumn{2}{|c|}{datasets} & classes  & samples  &   features\\
		\hline
		\hline
		\multirow{6}{*}{multi-class} &   LUNG  & 5  &203  &  3312\\
		& USPS  & 10  & 9258  &  256\\
		& COIL20  & 20  &1440  &  1024\\
		& UMIST &   20  &575  &  10304\\
		& YaleB   & 38  &2414  &  1024\\
		& Isolet1   & 26  &1560  &  617\\
		\hline
		\multirow{3}{*}{multi-class} & Barcelona  & 4  &129  &  384\\
		&MSRC & 23  &591  &  384\\
		&TRECVID  & 39  &3721  &  512\\
		\hline
	\end{tabular}
	\label{tbl:datasets}
\end{table}

On each multi-class dataset, our method CSFS is compared to the several popular multi-class feature selection methods, which can be summarized as following:

\begin{itemize}
	\item \textbf{ReliefF} \cite{reliefF} selects features by finding the near-hit and near-miss instances using the $\ell_1$-norm.
	\item \textbf{Information Gain (IG)} \cite{InfoGain} selects features by using information gain as the split criterion in decision tree induction.
	\item \textbf{Minimum Redundancy Maximum Relevance (mRMR)} \cite{mRMR} selects features by calculating the feature relevance and redundancy according to their mutual information.
	\item \textbf{F-statistic} \cite{F-statistic} selects features by computing the scores for each feature according to between-group and within-group variabilities.
	\item \textbf{Robust Feature Selection (RFS)} \cite{nips2010rfs} selects features by joint $\ell_{2,1}$-norm minimization on both loss function and regularization.
\end{itemize}

For multi-label classification, the proposed method CSFS is compared with five representative multi-label feature selection methods on each multi-label dataset. These multi-label feature selection methods are summarized as following:

\begin{itemize}
	\item \textbf{Multi-Label ReliefF (MLReliefF)} \cite{cvpr2012MLR} improves the classical ReliefF by overcoming the ambiguity problem relating to multi-label classification.
	\item \textbf{Multi-Label F-statistic (MLF-statistic)} \cite{cvpr2012MLR} incorporates the multi-label information into the conventional F-statistic algorithm.
	\item \textbf{Information-Theoretic Feature Ranking (ITFR)} \cite{ITFR2015} selects features by avoiding and reusing entropy calculations and identifying important label combinations based on information theory.
	\item \textbf{Non-Convex Feature Learning (NCFS)} \cite{NCFS} selects features by using the proximal gradient method to solve the $\ell_{p,\infty}$ operator problem.
	\item \textbf{Robust Feature Selection (RFS)} \cite{nips2010rfs} can be naturally extended for multi-label feature selection tasks \cite{NCFS}.
\end{itemize}

In addition, we also compare CSFS with two binary-class feature selection methods which are designed for imbalanced data:

\begin{itemize}
	\item \textbf{FAST} \cite{SIGKDD2008fast} takes each feature as the output of a classifier, calculates AUC for each feature by sliding thresholds, and then selects features according to AUC results in a descending order.
	\item \textbf{FAIR} \cite{KDE2010combating} is a modification of FAST which chooses features according to their corresponding largest precision-recall curve values. 
\end{itemize}

In the training process, the parameter $\lambda$ in our method is optimized in the range of $\{10^{-6}, 10^{-5},\dots,10^6\}$, and the numbers of selected features are set as $\{20, 30, \dots,120\}$. To fairly compare all different feature selection methods, classification experiments are conducted on all datasets using 5-fold cross validation SVM with linear kernel and parameter $C=1$. We repeat the experiments 10 times with random seeds for generating the validation sets. Both mean and standard deviation of the accuracy and F$_1$-measures are reported.

\subsection{Synthetic Data}

To demonstrate the advantage of cost-sensitive feature selection over traditional cost-blind feature selection, a toy experiment is conducted to show the influence of the costs on the selected features. We construct a two-dimensional binary-class synthetic dataset based on two different uniform distributions, as shown in Figure \ref{fig:toyExp}. The ratio of majority class to minority class is $3:1$.  In this experiment, majority class is treated as the positive class, and minority class as the negative class.

As mentioned above, regularized regression-based feature selection methods achieve the classifier learning and feature selection procedures simultaneously.  For a given linear classifier,  each coefficient of its projection vector $\mathbf{w}$ corresponds to one feature weight such as $w_1$ for $x_1$, then the features with larger coefficients will be selected. The  projection vector $\mathbf{w}$ vary with the costs assigned to both classes. In Figure \ref{fig:toyExp}(a), the cost of majority class is larger than the cost of minority class when $r<1$.  In Figure \ref{fig:toyExp}(b), the costs for both classes are the same when $r=1$. In this case, the cost-sensitive feature selection degenerates to the cost-blind feature selection.  When $r>1$, as shown in Figure \ref{fig:toyExp}(c), the cost of majority class is smaller than the cost of minority class. It is worth noting that the weight of feature $x_1$ is larger than the weight of feature $x_2$, which is different from the first two examples. Therefore, different features will be selected from different cost-sensitive feature selection problems.

\subsection{Binary-Class Datasets with Imbalanced Samplings }

Re-sampling technique has been used to deal with the class imbalance problem by over-sampling the minority class and under-sampling the majority class \cite{sampling_elkan2001foundations}. However, the over/under-sampling processes both change the statistical distribution of original data. Besides, the sampling ratio is hard to determine for binary class datasets, and this problem becomes more difficult for real-world multi-class and multi-label datasets.

To evaluate the performance of our CSFS and other feature selection methods under varying conditions of class imbalance, we construct multiple binary-class datasets with different sampling ratios. These datasets  are extracted for a two-class subset of the USPS dataset. The number of samples for one class is fixed at 150, while the number of samples for the other class varies in the set $\{150,300,\dots,1500\}$. In this way, we obtain 10 binary-class datasets with different levels of class imbalance.

Figure \ref{fig:imbalanceFmAcc} shows the classification F$_1$-measure and accuracy comparisons of top 20, 70 and 120 features selected by ReliefF, IG, mRMR, F-statistic, RFS and the proposed CSFS on these ten datasets.
Moreover, we also add a baseline named CSF-SVM. It first trains cost-sensitive SVMs on the training set within the framework of \cite{nips2014optimizing}, then slides the thresholds from the minimum to the maximum prediction scores of the validation samples with the step of 0.01 to obtain the optimal F-measure classifier, and finally uses the optimal classifier to perform feature selection on test set after sorting the feature weights.
This baseline method can be used to show how the proposed CSFS significantly improves the quality of selected features compared to the traditional cost-sensitive classifiers which are used for classification \cite{nips2014optimizing}. These results show that: (1) as the proportion of the two classes becomes larger, all the compared feature selection methods except the baseline achieve similar accuracy performance. This is because when the classes are imbalanced, accuracy is not an appropriate performance measure; (2) the proposed CSFS outperforms the other methods significantly under the F$_1$-measure criterion. The performances of the others decrease sharply by increasing the proportion, while the curve of CSFS is steady; and (3) although the F$_1$-measure of CSF-SVM is relatively lower, its F$_1$-measure descends slowly as the ratio becomes larger. This indicates that the introduction of optimizing F-measure through cost-sensitive classification benefits feature selection.

To obtain a more comprehensive understanding of the performance of our method, we compare CSFS with two competitive algorithms which take class imbalance into account, FAST \cite{SIGKDD2008fast} and FAIR \cite{KDE2010combating}. Since FAST and FAIR are only designed for binary-class learning tasks, we conduct the comparison experiment in terms of F$_1$-measure, accuracy and AUC on the USPS 1:10 subset. The classification results are shown in Figure \ref{fig_binaryFastFairCsfs}. It can be seen that: (1) CSFS achieves similar AUC performances compared with FAST and FAIR when more than 80 features are selected; and (2) CSFS outperforms FAST and FAIR significantly on F$_1$-measure, and demonstrates the consistent superiority on accuracy. These results can be explained from two aspects. On one hand, both F-measure used on CSFS as well as AUC used in FAST and FAIR are appropriate measures for imbalanced datasets, and thus they display advantages over other methods that neglect the imbalance issue. On the other hand, FAST and FAIR independently select each individual feature and AUC is not involved in the optimization procedure. As a result, their superiority on AUC gradually drops as the number of selected features increases.

\subsection{Comparisons with Multi-Class Feature Selection Methods}
To compare the proposed algorithm with other multi-class feature selection methods, we conduct our experiments on multi-class datasets LUNG, USPS, COIL20, UMIST, YaleB and Isolet1. On each dataset, our method CSFS is compared to five multi-class feature selection methods, such as ReliefF \cite{reliefF}, Information Gain (IG) \cite{InfoGain}, mRMR \cite{mRMR}, F-statistic \cite{F-statistic} and RFS \cite{nips2010rfs}.

We follow the experimental setups of the previous works \cite{cvpr2015ufs,nips2010rfs}, and employ classification accuracy and multi-class micro-F$_1$-measure to evaluate the classification results of each feature set selected by various methods. 

\begin{figure*}[!tbp]
\centering
\includegraphics[width=0.92\textwidth]{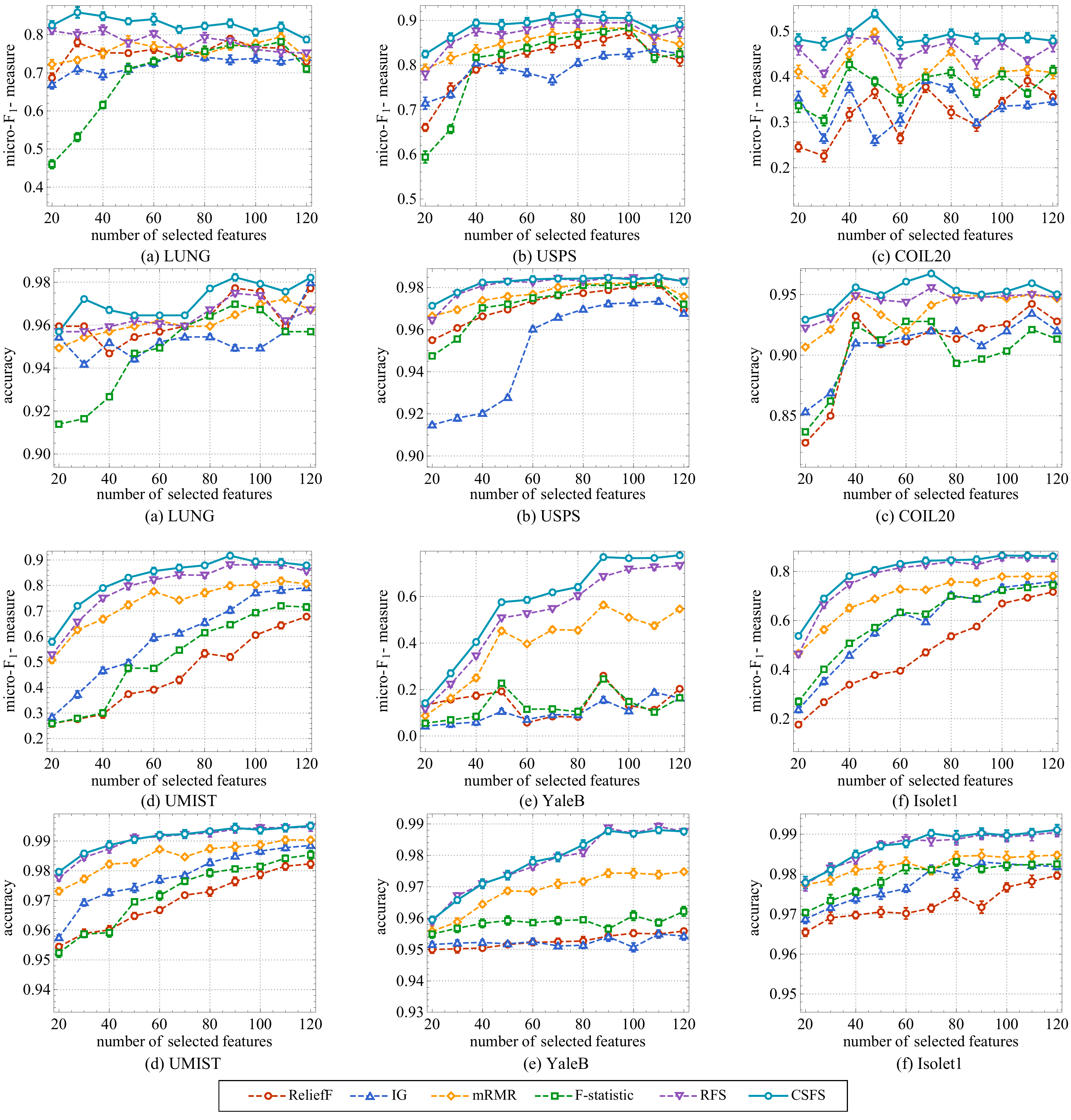}\\
\caption{Multi-class classification results using SVM in terms of multi-class micro-F$_1$-measure and accuracy on LUNG, USPS, COIL20, UMIST, YaleB and Isolet1 datasets. The proposed CSFS is compared with five multi-class feature selection methods, such as ReliefF, IG, mRMR, F-statistic and RFS. For each dataset, the chart above shows the multi-class micro-F$_1$-measure and the chart below shows the accuracy of classification.}
\label{fig:multiClassFmAcc}
\end{figure*}

Figure \ref{fig:multiClassFmAcc} shows the multi-class classification results in terms of micro-F$_1$-measure and accuracy. Table \ref{tbl:multiClassFmAcc}  shows the results of different feature selection methods on their best dimensions. We observe that: (1) the proposed CSFS is superior to other multi-class feature selection methods consistently in terms of the micro-F$_1$-measure on all six datasets; (2) in terms of accuracy, CSFS outperforms other methods on most of the selected features on all six datasets. Besides, we can also note that compared with CSFS, RFS is quite competitive on spoken letter recognition dataset Isolet1. This can be explained from two aspects. On one hand, CSFS and RFS both use $\ell_{2,1}$-norm on loss term and regularization term. When setting the discretized F-measure value as 1, the misclassification costs are equal and the cost-sensitive CSFS degenerates to the cost-blind RFS. On the other hand, we only set the number of discretized values of F-measure $T$ as 20. The performance of CSFS can be further improved by using smaller discretized steps.  Therefore, the competitivity of RFS validates rather than decreases the effectiveness of CSFS.

\begin{table*}[!tbp]
\setlength\tabcolsep{12.5pt}
\renewcommand{\arraystretch}{1.2}
\centering
\caption{Multi-class micro-F$_1$-measure ($\%\pm$ std) and accuracy ($\%\pm$ std) of various multi-class feature selection methods on six datasets. The best results are in boldface.}
\begin{tabular}{|c||c|c|c|c|c|c|}
  \hline

        & LUNG & USPS & COIL20 & UMIST & YaleB & Isolet1    \\
  \hline
    \hline
    & \multicolumn{6}{c|}{Micro-F$_1$-measure}   \\

  \hline

  ReliefF    & 78.02$\pm$7.35 & 87.25$\pm$0.71 & 67.78$\pm$4.47 & 25.94$\pm$6.95 &  39.09$\pm$6.55  & 71.65$\pm$1.21 \\
	 	 	 	 	 	 	 	 	
  IG  & 74.99$\pm$5.87 & 83.51$\pm$1.06 & 79.13$\pm$2.86 & 18.78$\pm$8.90 &  39.13$\pm$1.47  & 75.89$\pm$0.90  \\
	 	 	 	 	 	 	 	
  mRMR  & 79.46$\pm$6.03 & 88.30$\pm$0.87 & 81.83$\pm$1.15 & 56.38$\pm$1.60 &  49.80$\pm$9.44  & 77.96$\pm$1.19  \\

  F-statistic  & 78.18$\pm$7.07 & 88.36$\pm$0.84 & 72.00$\pm$3.31 & 24.53$\pm$2.41 &  42.64$\pm$1.13  & 74.45$\pm$2.40  \\

  RFS  & 81.29$\pm$9.33 & 89.54$\pm$0.62 & 88.15$\pm$2.00 & 73.33$\pm$4.18 &  48.68$\pm$8.54  & 85.43$\pm$0.94  \\

  CSFS & \textbf{85.88}$\pm$\textbf{4.70} & \textbf{91.56}$\pm$\textbf{0.56} & \textbf{91.69}$\pm$\textbf{1.19} & \textbf{77.77}$\pm$\textbf{1.50} &  \textbf{53.83}$\pm$\textbf{1.63}  & \textbf{87.49}$\pm$\textbf{0.81}  \\
  \hline
    \hline
    & \multicolumn{6}{c|}{Accuracy}   \\
  \hline
  ReliefF    & 97.72$\pm$1.02 & 98.13$\pm$1.39 & 98.23$\pm$1.30 & 95.59$\pm$0.41 &  94.22$\pm$0.34  & 97.97$\pm$0.12 \\

  IG  & 97.97$\pm$0.56 & 97.35$\pm$2.96 & 98.85$\pm$3.66 & 95.41$\pm$2.06 &  93.44$\pm$1.22  & 98.31$\pm$0.28 \\

  mRMR  & 97.22$\pm$0.55 & 98.23$\pm$0.76 & 99.04$\pm$1.16 & 97.48$\pm$0.20 &  95.00$\pm$0.85  & 98.27$\pm$0.13\\

  F-statistic  & 96.97$\pm$1.64 & 98.20$\pm$1.69 & 98.54$\pm$1.84 & 96.22$\pm$0.24 &  92.78$\pm$2.70  & 98.25$\pm$0.07  \\

  RFS  & 97.48$\pm$1.45 & 98.44$\pm$0.95 & 99.46$\pm$1.43 & 98.62$\pm$0.15 &  95.56$\pm$1.51  & 99.04$\pm$0.05 \\

  CSFS & \textbf{98.23}$\pm$\textbf{0.45} & \textbf{98.50}$\pm$\textbf{0.56} & \textbf{99.51}$\pm$\textbf{1.04} & \textbf{98.82}$\pm$\textbf{0.14} &  \textbf{96.72}$\pm$\textbf{0.41}  & \textbf{99.11}$\pm$\textbf{0.07} \\
  \hline
\end{tabular}
\label{tbl:multiClassFmAcc}
\end{table*}

\begin{figure*}[!tbp]
	\centering
	\includegraphics[width=0.92\textwidth]{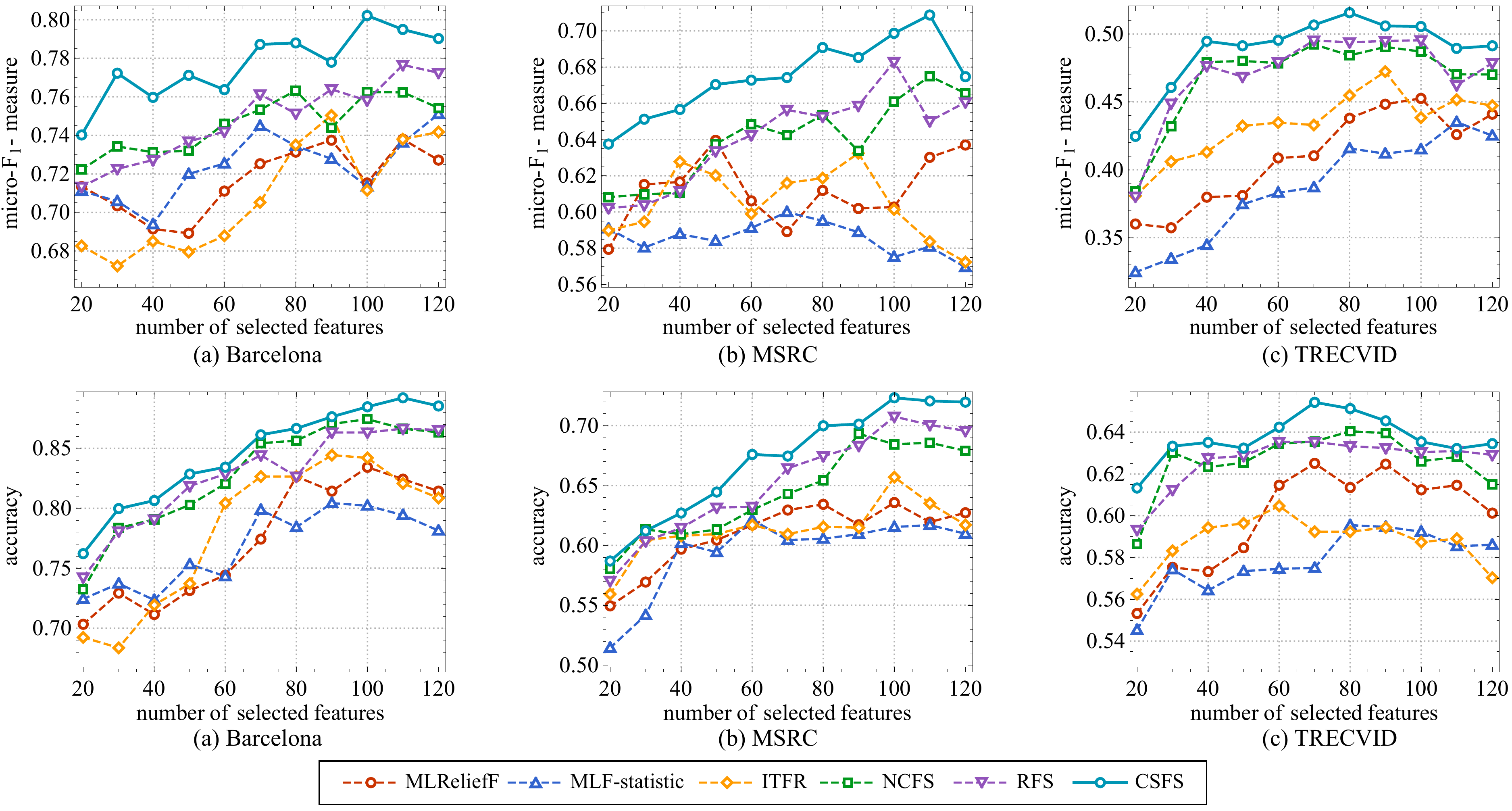}\\
	\caption{Multi-label classification results using SVM in terms of multi-label micro-F$_1$-measure and accuracy on Barcelona, MSRC and TRECVID datasets. The proposed CSFS is compared with five multi-label feature selection methods, such as MLReliefF, MLF-statistic, ITFR, NCFS and RFS. For each dataset, the chart above shows the multi-label micro-F$_1$-measure and the chart below shows the accuracy of classification.}
	\label{fig:multilabelFmAcc}
\end{figure*}

\begin{table*}[!tbp]
\setlength\tabcolsep{12pt}
\renewcommand{\arraystretch}{1.2}
\centering
\caption{Multi-label micro-F$_1$-measure ($\%\pm$ std) and accuracy ($\%\pm$ std) of various multi-label feature selection methods on three datasets. The best results are in boldface.}
\begin{tabular}{|c||c|c|c||c|c|c|}

  \hline
    & Barcelona & MSRC & TRECVID & Barcelona & MSRC & TRECVID   \\
  \hline
 \hline
 & \multicolumn{3}{c||}{Micro-F$_1$-measure} & \multicolumn{3}{c|}{Accuracy}   \\
 \hline
  MLReliefF    & 73.80$\pm$5.70 & 63.96$\pm$0.49 & 45.25$\pm$0.71 & 83.42$\pm$8.42 &  63.57$\pm$1.41  & 62.46$\pm$0.88 \\

  MLF-statistic   & 75.10$\pm$3.61 & 59.99$\pm$1.87 & 43.51$\pm$0.36 & 80.43$\pm$0.89 &  62.15$\pm$1.50  & 59.55$\pm$4.27 \\

  ITFR & 75.03$\pm$4.32 & 63.24$\pm$1.07 & 47.23$\pm$0.59 & 84.43$\pm$2.25 &  65.70$\pm$0.32  & 60.46$\pm$3.75\\

  NCFS  & 76.32$\pm$5.08 & 67.50$\pm$1.96 & 49.23$\pm$0.80 & 87.43$\pm$1.18 &  69.32$\pm$1.64  & 64.04$\pm$3.57  \\

  RFS  & 77.65$\pm$2.70 & 68.29$\pm$0.93 & 49.54$\pm$0.62 & 86.64$\pm$3.05 &  70.75$\pm$1.02  & 63.53$\pm$0.77 \\

  CSFS & \textbf{80.22}$\pm$\textbf{0.91} & \textbf{70.88}$\pm$\textbf{0.77} & \textbf{51.56}$\pm$\textbf{0.56} & \textbf{89.21}$\pm$\textbf{1.05} &  \textbf{72.32}$\pm$\textbf{0.46}  & \textbf{65.42}$\pm$\textbf{0.43} \\
  \hline
\end{tabular}
\label{tbl:multiLabelFmAcc}
\end{table*}

\subsection{Comparisons with Multi-Label Feature Selection Methods}

To compare our CSFS with other multi-label feature selection methods, we conduct the experiments on multi-label datasets Barcelona, MSRC and TREVCID by following the setups of the previous works \cite{nips2010rfs,NCFS,cvpr2012MLR}. On each multi-label dataset, the proposed CSFS is compared with five competitive multi-label feature selection methods: MLReliefF \cite{cvpr2012MLR}, MLF-statistic \cite{cvpr2012MLR}, ITFR \cite{ITFR2015}, NCFS \cite{NCFS} and RFS \cite{nips2010rfs}.

Figure \ref{fig:multilabelFmAcc} shows the multi-label classification results in terms of micro-F$_1$-measure and accuracy on Barcelona, MSRC and TRECVID datasets. Table \ref{tbl:multiLabelFmAcc} shows the results of each feature selection method on its best performing dimension. From the results, we observe that: (1) the methods using joint sparse regularization, such as CSFS, NCFS and RFS, show better performances than other feature selection methods that only use the statistical information of the original features. This is because the projection matrices of these methods are determined at the same time during the optimization procedure, corresponding features are selected to prevent high correlation \cite{cvpr2015ufs}; (2) the feature selection methods based on accuracy optimization have poor F-measure performance because they tend to select those features that are biased towards the majority class. Our proposed method outperforms these methods significantly under the F-measure criterion, and does not lead to obvious decrement on accuracy. In particular, our method outperforms other methods by a relative improvement between 2$\%$-$10\%$ on all three datasets in terms of micro-F$_1$-measure.

\section{Conclusion}

In this paper, we present a novel feature selection method by optimizing F-measure instead of accuracy to tackle the class imbalance problem. Due to the neglect of class imbalance issue, conventional feature selection methods usually select the features that are biased towards the majority class. For instance, embedded methods select the feature subset by maximizing the classification accuracy, which is unsuitable given the imbalanced classes. In this situation, F-measure is a more appropriate performance measure than accuracy.  Based on the reduction of F-measure optimization to cost-sensitive classification, we modify the classifiers of regularized regression-based feature selection into cost-sensitive by generating and assigning different costs to each class. After solving a series of cost-sensitive feature selection problems, features will be selected according to the classifier with optimal F-measure. Thus the selected features will be more representative for all classes. Extensive experiments on both multi-class and multi-label datasets demonstrate the effectiveness of the proposed method.


%


\appendices
\section{Proof of Theorem 1}
\begin{proof}

	In each iteration $t$, optimal $\mathbf{W}_{t+1}$ is obtained by
	\begin{align}
	&\mathbf{W}_{t+1} = \nonumber\\
	&\min_{\textbf{W}}Tr(((\mathbf{X}^T\mathbf{W}-\mathbf{Y})\odot \mathbf{C})^T\mathbf{G}_{t+1}((\mathbf{X}^T\mathbf{W}-\mathbf{Y})\odot \mathbf{C}))\nonumber\\
	&+\lambda Tr(\mathbf{W}^T\mathbf{D}_{t+1}\mathbf{W}\label{eqn:optW_eachIteration}).
	\end{align}
	
	Fixing $\mathbf{D}_{t+1}$ and $\mathbf{G}_{t+1}$, we have:
	\begin{align}
	&Tr\displaystyle(((\mathbf{X}^T\mathbf{W}_{t+1}-\mathbf{Y})\odot \mathbf{C})^T\mathbf{G}_{t+1}((\mathbf{X}^T\mathbf{W}_{t+1}-\mathbf{Y})\odot \mathbf{C})\displaystyle)\nonumber\\
	&+\lambda Tr(\mathbf{W}_{t+1}^T\mathbf{D}_{t+1}\mathbf{W}_{t+1})\nonumber\\
	&\leq  Tr\displaystyle(((\mathbf{X}^T\mathbf{W}_{t}-\mathbf{Y})\odot \mathbf{C})^T\mathbf{G}_{t+1}((\mathbf{X}^T\mathbf{W}_{t}-\mathbf{Y})\odot \mathbf{C}))\nonumber\\
	&+\lambda Tr(\mathbf{W}_{t}^T\mathbf{D}_{t+1}\mathbf{W}_{t}).
	\end{align}
	
	Substituting diagonal matrices $\mathbf{D}_{t+1}$ and $\mathbf{G}_{t+1}$ with their corresponding definitions, we obtain:
	\begin{align}
	\label{eqn:convergence_Main}
	&\sum_{i=1}^n\frac{\|((\mathbf{X}^T\mathbf{W}_{t+1}-\mathbf{Y})\odot \mathbf{C})^i\|_2^2}{2\|((\mathbf{X}^T\mathbf{W}_{t}-\mathbf{Y})\odot \mathbf{C})^i\|_2}
	+\lambda\sum_{i=1}^d\frac{\|\mathbf{w}^i_{t+1}\|_2^2}{2\|\mathbf{w}^i_{t}\|_2}\leq\nonumber\\
	& \sum_{i=1}^n\frac{\|((\mathbf{X}^T\mathbf{W}_{t}-\mathbf{Y})\odot \mathbf{C})^i\|_2^2}{2\|((\mathbf{X}^T\mathbf{W}_{t}-\mathbf{Y})\odot \mathbf{C})^i\|_2}+\lambda\sum_{i=1}^d\frac{\|\mathbf{w}^i_{t}\|_2^2}{2\|\mathbf{w}^i_{t}\|_2}.
	\end{align}
	
	Because it can be easily verified that function $f(x)=x-\frac{x^2}{2a}$ achieves its maximum value when $a=x$, then we have
	\begin{equation}\label{eqn:convergence_Support_1}
	\begin{split}
	&\displaystyle\sum_{i=1}^n\|((\mathbf{X}^T\mathbf{W}_{t+1}-\mathbf{Y})\odot \mathbf{C})^i\|_2-\\
	&\displaystyle\sum_{i=1}^n\frac{\|((\mathbf{X}^T\mathbf{W}_{t+1}-\mathbf{Y})\odot \mathbf{C})^i\|_2^2}
	{2\|((\mathbf{X}^T\mathbf{W}_{t}-\mathbf{Y})\odot \mathbf{C})^i\|_2}\leq\\
	&\displaystyle
	\sum_{i=1}^n\|((\mathbf{X}^T\mathbf{W}_{t}-\mathbf{Y})\odot \mathbf{C})^i\|_2-\\
	&\displaystyle\sum_{i=1}^n\frac{\|((\mathbf{X}^T\mathbf{W}_{t} -\mathbf{Y})\odot \mathbf{C})^i\|_2^2}
	{2\|((\mathbf{X}^T\mathbf{W}_{t}-\mathbf{Y})\odot \mathbf{C})^i\|_2},
	\end{split}
	\end{equation}
	and
	\begin{equation}\label{eqn:convergence_Support_2}
	\begin{split}
	&\sum_{i=1}^d\|\mathbf{w}^i_{t+1}\|_2-\sum_{i=1}^d\frac{\|\mathbf{w}^i_{t+1}\|_2^2}{2\|\mathbf{w}^i_{t}\|_2}\leq\\
	& \sum_{i=1}^d\|\mathbf{w}^i_{t}\|_2-\sum_{i=1}^d\frac{\|\mathbf{w}^i_{t}\|_2^2}{2\|\mathbf{w}^i_{t}\|_2}.
	\end{split}
	\end{equation}

	Adding Eqs. (\ref{eqn:convergence_Main}-\ref{eqn:convergence_Support_2}) on both sides, we get
	\begin{align}
	&\displaystyle\sum_{i=1}^n\|((\mathbf{X}^T\mathbf{W}_{t+1}-\mathbf{Y})\odot \mathbf{C})^i\|_2+\lambda\sum_{i=1}^d\|\mathbf{w}^i_{t+1}\|_2^2\leq\nonumber\\
	&\displaystyle \sum_{i=1}^n\|((\mathbf{X}^T\mathbf{W}_{t}-\mathbf{Y})\odot \mathbf{C})^i\|_2+\lambda\sum_{i=1}^d\|\mathbf{w}^i_{t}\|_2^2,
	\end{align}
	that is,
	\begin{equation}
	\begin{split}
	&\|(\mathbf{X}^T\mathbf{W}_{t+1}-\mathbf{Y})\odot \mathbf{C}\|_{2,1} +\lambda\|\mathbf{W}_{t+1}\|_{2,1}\\
	&\leq\|(\mathbf{X}^T\mathbf{W}_{t}-\mathbf{Y})\odot \mathbf{C}\|_{2,1} +\lambda\|\mathbf{W}_{t}\|_{2,1}.
	\end{split}
	\end{equation}
	
	Therefore, Algorithm \ref{alg:CSFS} decreases the objective value monotonically in each iteration.
\end{proof}

%

\section*{Acknowledgements}
The authors would like to thank the handling associate editor Dr. Paul Rodriguez and both anonymous reviewers for their constructive comments.
This research is partially supported by NSFC under Grant 61375026 and Grant 2015BAF15B00, Australian Research Council Projects FL-170100117, DP-180103424, DP-140102164, LP-150100671, DE-180101438, and Singapore MOE Tier 1 projects (RG 17/14, RG 26/16) and NRF2015ENC-GDCR01001-003 (administrated via IMDA).

\ifCLASSOPTIONcaptionsoff
  \newpage
\fi



%
\bibliographystyle{IEEEtran}
\bibliography{csfsBIB}

%
%

%



\begin{IEEEbiography}[{\includegraphics[width=1in,height=1.25in,clip,keepaspectratio]{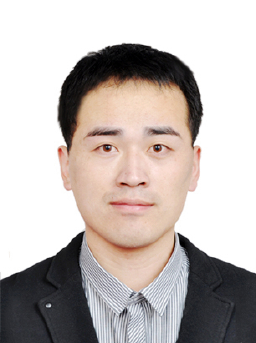}}]{Meng Liu}
	received the B.S. degree in 2012 and the D.Sc degree in 2017 from School of Electronics Engineering and Computer Science, Peking University, Beijing, China. He is currently working as a research engineer in Laboratory 2012 of Huawei Corporation, Beijing, China. He was a Visiting Scholar with the Faculty of Engineering and Information Technology, University of Technology, Sydney, Australia. His current research interests include machine learning and its applications on classification and imbalance issues.
\end{IEEEbiography}


%
\begin{IEEEbiography}[{\includegraphics[width=1in,height=1.25in,clip,keepaspectratio]{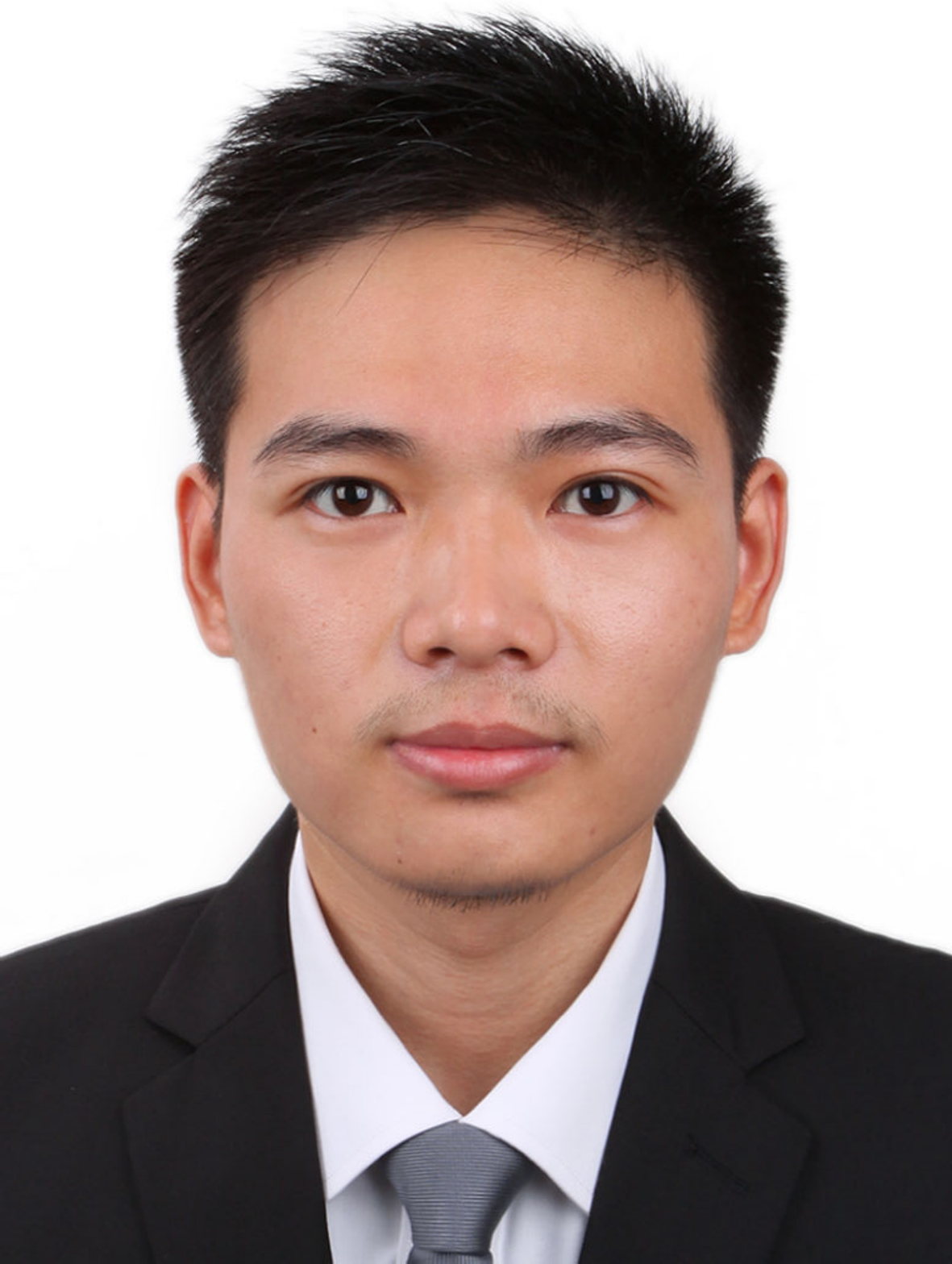}}]{Chang Xu}
	is a Lecturer in Machine Learning and Computer Vision at the School of Information Technologies, University of Sydney. He obtained the B.E. degree from Tianjin University, China, and the Ph.D. degree from Peking University, China. While pursing the PhD degree, he received the PhD fellowships from IBM and Baidu. His research interests lie in Machine Learning algorithms and related applications in Computer Vision, including multi-view learning, multi-label learning, and visual search. 
\end{IEEEbiography}


\begin{IEEEbiography}[{\includegraphics[width=1in,height=1.25in,clip,keepaspectratio]{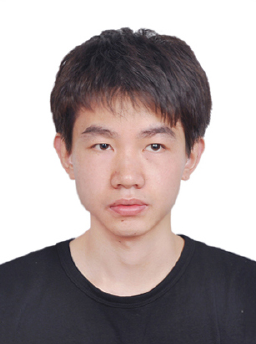}}]{Yong Luo}
	received the B.E. degree in Computer Science from the Northwestern Polytechnical University, Xi'an, China, in 2009, and the D.Sc. degree in the School of Electronics Engineering and Computer Science, Peking University, Beijing, China, in 2014. He is currently a Research Fellow with the School of Computer Science and Engineering, Nanyang Technological University. He was a visiting student in the School of Computer Engineering, Nanyang Technological University, and the Faculty of Engineering and Information Technology, University of Technology Sydney. His research interests are primarily on machine learning and data mining with applications to visual information understanding and analysis. He has authored several scientific articles at top venues including IEEE T-NNLS, IEEE T-IP, IEEE T-KDE, IJCAI and AAAI. He received the IEEE Globecom 2016 Best Paper Award, and was nominated as the IJCAI 2017 Distinguished Best Paper Award.
\end{IEEEbiography}


\begin{IEEEbiography}[{\includegraphics[width=1in,height=1.25in,clip,keepaspectratio]{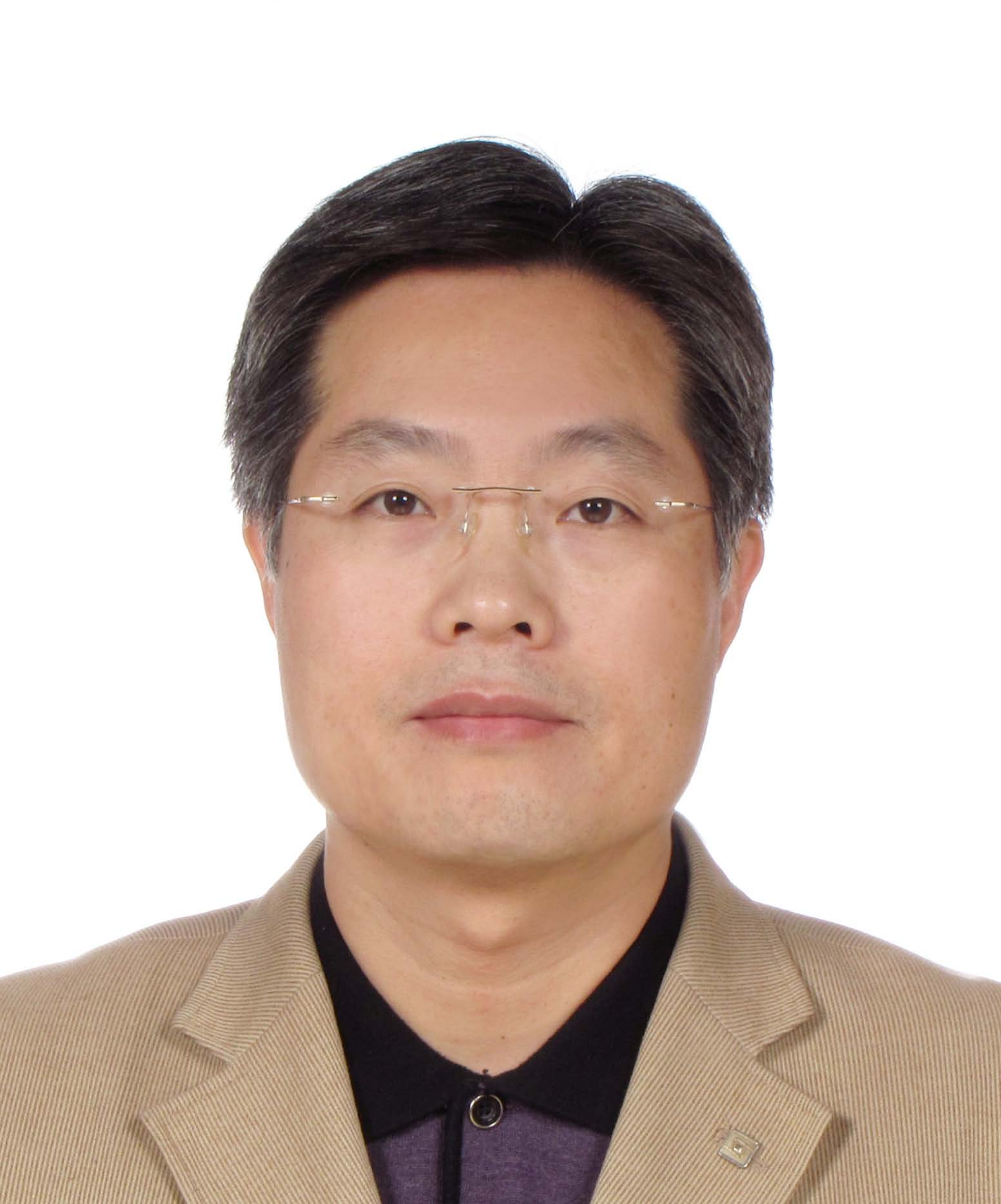}}]{Chao Xu}
	received the B.E. degree from Tsinghua University, Beijing, China, the M.S. degree from the University of Science and Technology of China, Hefei, China, and the Ph.D. degree from the Institute of Electronics, Chinese Academy of Sciences, Beijing, in 1988, 1991, and 1997, respectively. 
	He was an Assistant Researcher with the University of Science and Technology of China from 1991 and 1994. Since 1997, he has been with the Key Laboratory of Machine Perception (Ministry of Education), Peking University, Beijing, where he has been a Professor since 2005. He has authored or co-authored more than 100 papers in journals and conferences, and holds six patents. His current research interests include image and video processing, multimedia technology.
\end{IEEEbiography}

\begin{IEEEbiography}[{\includegraphics[width=1in,height=1.25in,clip,keepaspectratio]{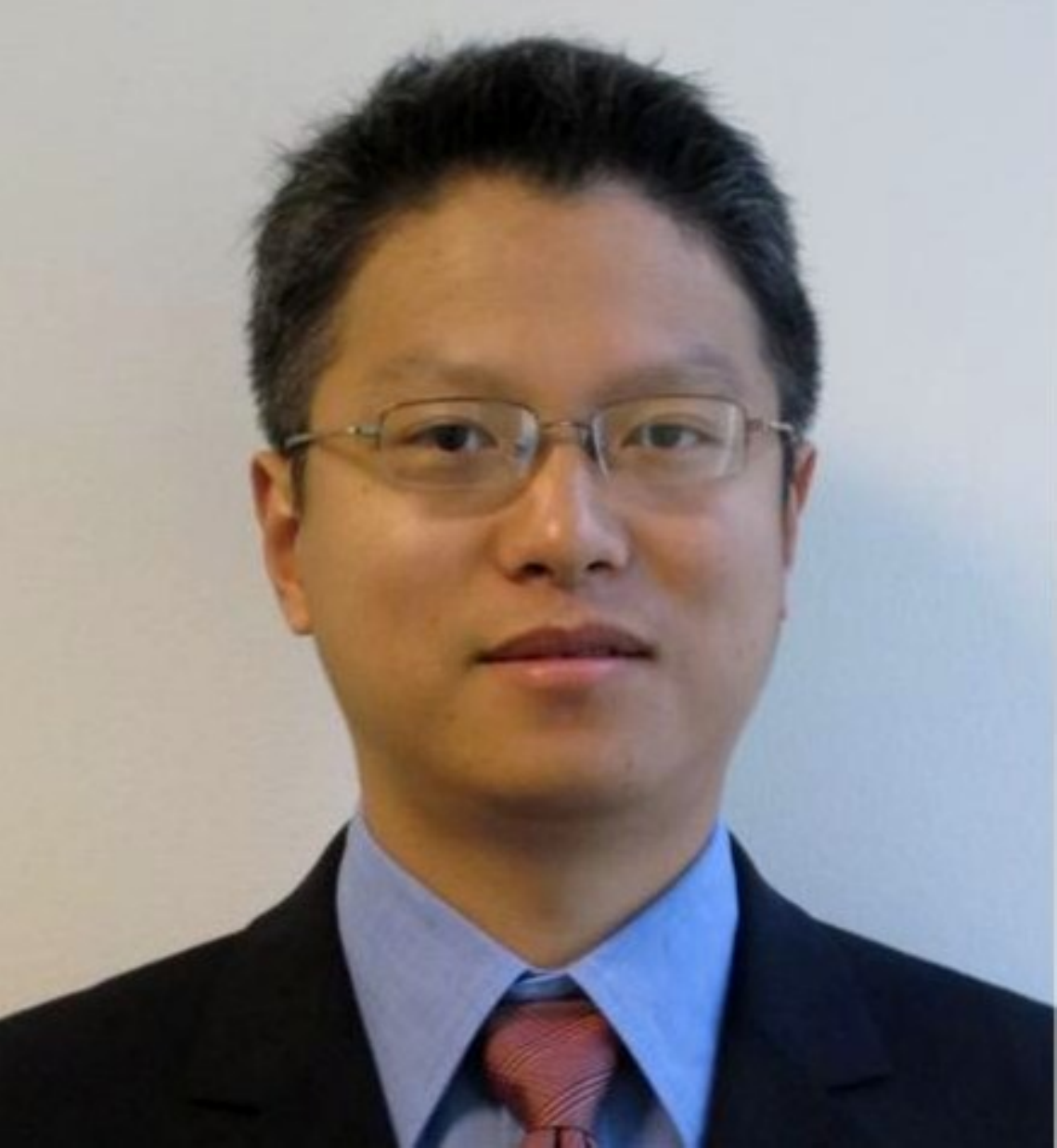}}]{Yonggang Wen}
	(S'99-M'08-SM'14) received the Ph.D. degree in electrical engineering and computer science (with a minor in western literature) from the Massachusetts Institute of Technology, Cambridge, MA, USA, in 2008. He is currently an Associate Professor with the School of Computer Science and Engineering, Nanyang Technological University, Singapore. He has been with Cisco, San Jose, CA, USA, where he led product development in content delivery network, which had a revenue impact of \$3 billion globally. His work in multiscreen cloud social TV has been featured by global media (over 1600 news articles from over 29 countries). He has authored or co-authored over 140 papers in top journals and prestigious conferences. His current research interests include cloud computing, green data centers, big data analytics, multimedia networks, and mobile computing. Dr. Wen was a recipient of the ASEAN ICT Award 2013 (Gold Medal) and the Data Center Dynamics Awards 2015-APAC for his work on cloud 3-D view, as the only academia entry, and a co-recipient of the 2015 IEEE Multimedia Best Paper Award, and the Best Paper Award at EAI/ICST Chinacom 2015, IEEE WCSP 2014, IEEE Globecom 2013, and IEEE EUC 2012. He was the Chair for the IEEE ComSoc Multimedia Communication Technical Committee (2014-2016).
\end{IEEEbiography}


\begin{IEEEbiography}[{\includegraphics[width=1in,height=1.25in,clip,keepaspectratio]{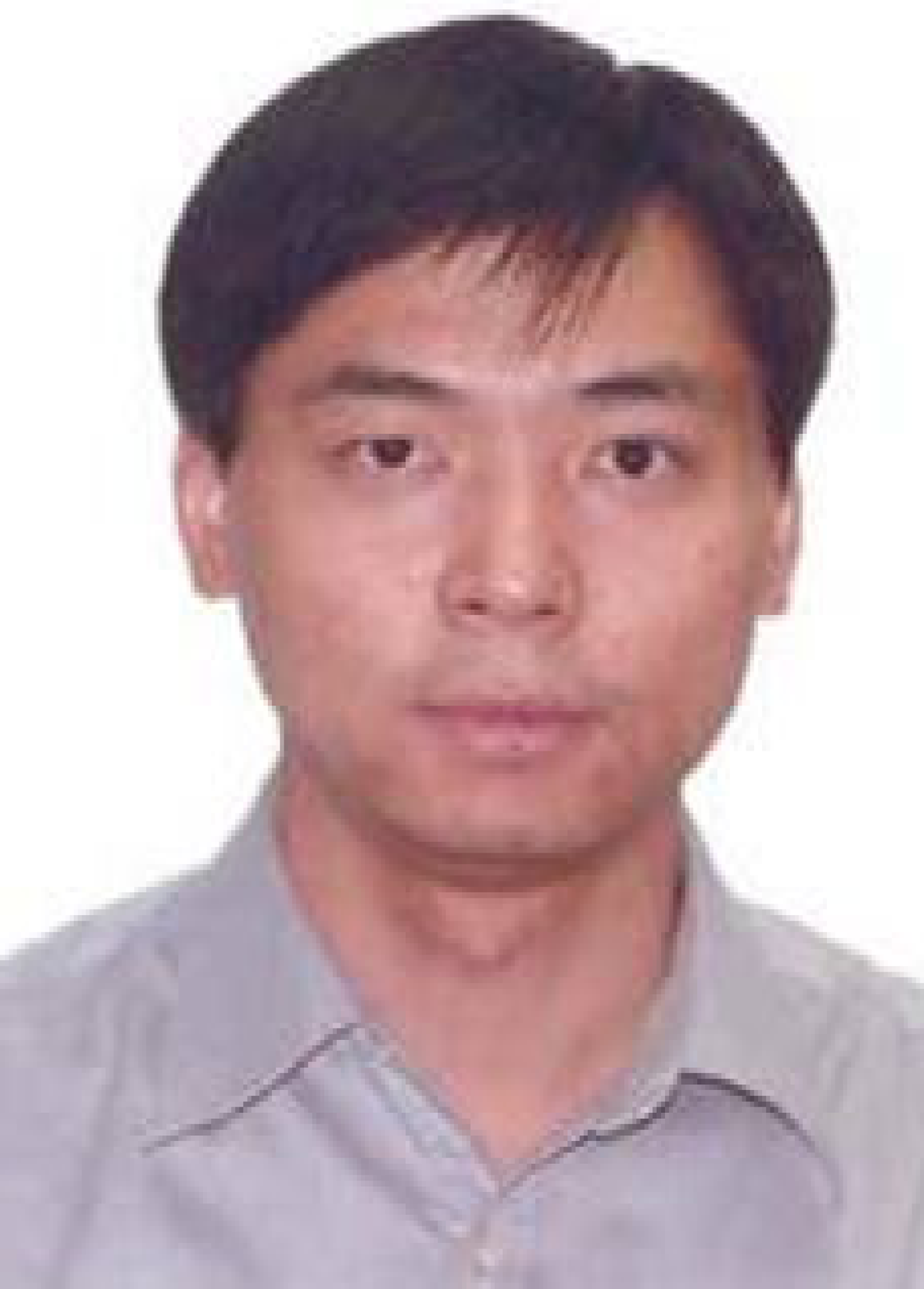}}]{Dacheng Tao}
	
	(F'15) is Professor of Computer Science and ARC Laureate Fellow in the School of Information Technologies and the Faculty of Engineering and Information Technologies, and the Inaugural Director of the UBTECH Sydney Artificial Intelligence Centre, at the University of Sydney. He mainly applies statistics and mathematics to Artificial Intelligence and Data Science. His research interests spread across computer vision, data science, image processing, machine learning, and video surveillance. His research results have expounded in one monograph and 500+ publications at prestigious journals and prominent conferences, such as IEEE T-PAMI, T-NNLS, T-IP, JMLR, IJCV, NIPS, ICML, CVPR, ICCV, ECCV, ICDM; and ACM SIGKDD, with several best paper awards, such as the best theory/algorithm paper runner up award in IEEE ICDM'07, the best student paper award in IEEE ICDM'13, the distinguished student paper award in the 2017 IJCAI, the 2014 ICDM 10-year highest-impact paper award, and the 2017 IEEE Signal Processing Society Best Paper Award. He received the 2015 Australian Scopus-Eureka Prize, the 2015 ACS Gold Disruptor Award and the 2015 UTS Vice-Chancellor's Medal for Exceptional Research. He is a Fellow of the IEEE, AAAS, OSA, IAPR and SPIE. 
\end{IEEEbiography}






\end{document}